\documentclass[sigconf]{acmart}

\usepackage{balance}

\def\BibTeX{{\rm B\kern-.05em{\sc i\kern-.025em b}\kern-.08emT\kern-.1667em\lower.7ex\hbox{E}\kern-.125emX}}

\usepackage{booktabs} 
\usepackage{amsmath}
\usepackage{algorithm}  
\usepackage{algorithmic} 
\usepackage{subfigure}
\usepackage{multirow}
\usepackage{tablefootnote}
\usepackage{enumitem}
\usepackage{bm}
    
\copyrightyear{2019} 
\acmYear{2019} 
\setcopyright{acmcopyright}
\acmConference[KDD '19]{The 25th ACM SIGKDD Conference on Knowledge Discovery and Data Mining}{August 4--8, 2019}{Anchorage, AK, USA}
\acmBooktitle{The 25th ACM SIGKDD Conference on Knowledge Discovery and Data Mining (KDD '19), August 4--8, 2019, Anchorage, AK, USA}
\acmPrice{15.00}
\acmDOI{10.1145/3292500.3330880}
\acmISBN{978-1-4503-6201-6/19/08}

\begin{document}

\title{\textsc{$\lambda$Opt}: Learn to Regularize Recommender Models in Finer Levels}


\author{Yihong Chen}
\authornote{This work was partly done when the author was at Microsoft Research.}
\affiliation{
  \institution{Dep. of Electr. Engin., Tsinghua Univ.}
}
\email{yihong-chen@outlook.com}

\author{Bei Chen}
\affiliation{
  \institution{Microsoft Research, Beijing, China}
}
\email{beichen@microsoft.com}

\author{Xiangnan He}
\affiliation{
  \institution{Univ. of Sci. and Technol. of China}
}
\email{xiangnanhe@gmail.com}

\author{Chen Gao}
\affiliation{
  \institution{Dep. of Electr. Engin., Tsinghua Univ.}
}
\email{gc16@mails.tsinghua.edu.cn}

\author{Yong Li}
\authornote{corresponding author}
\affiliation{
  \institution{Dep. of Electr. Engin., Tsinghua Univ.}
}
\email{liyong07@tsinghua.edu.cn}

\author{Jian-Guang Lou}
\affiliation{
  \institution{Microsoft Research, Beijing, China}
}
\email{jlou@microsoft.com}

\author{Yue Wang}
\affiliation{
  \institution{Dep. of Electr. Engin., Tsinghua Univ.}
}
\email{wangyue@mail.tsinghua.edu.cn}

%
\renewcommand{\shortauthors}{Y. Chen, B. Chen, X. He, C. Gao, Y. Li, J. Lou, and Y. Wang}

%
\begin{abstract}
Recommendation models mainly deal with categorical variables, such as user/item ID and attributes. Besides the high-cardinality issue, the interactions among such categorical variables are usually long-tailed, with the head made up of highly frequent values and a long tail of rare ones. This phenomenon results in the data sparsity issue, making it essential to regularize the models to ensure generalization. The common practice is to employ grid search to manually tune regularization hyperparameters based on the validation data. However, it requires non-trivial efforts and large computation resources to search the whole candidate space; even so, it may not lead to the optimal choice, for which different parameters should have different regularization strengths. 

In this paper, we propose a hyperparameter optimization method, \textsc{$\lambda$Opt}\footnote{Codes can be found on https://github.com/LaceyChen17/lambda-opt.}, which automatically and adaptively enforces regularization during training. Specifically, it updates the regularization coefficients based on the performance of validation data. With \textsc{$\lambda$Opt}, the notorious tuning of regularization hyperparameters can be avoided; more importantly, it allows fine-grained regularization (i.e. each parameter can have an individualized regularization coefficient), leading to better generalized models. We show how to employ \textsc{$\lambda$Opt} on matrix factorization, a classical model that is representative of a large family of recommender models. Extensive experiments on two public benchmarks demonstrate the superiority of our method in boosting the performance of top-K recommendation. 
\end{abstract}

%
%


%

%

%
\maketitle

\section{Introduction}\label{sec:introduction}
\begin{figure}[h]
    \centering
    \vspace{-1mm}
    \includegraphics[width=0.85\linewidth]{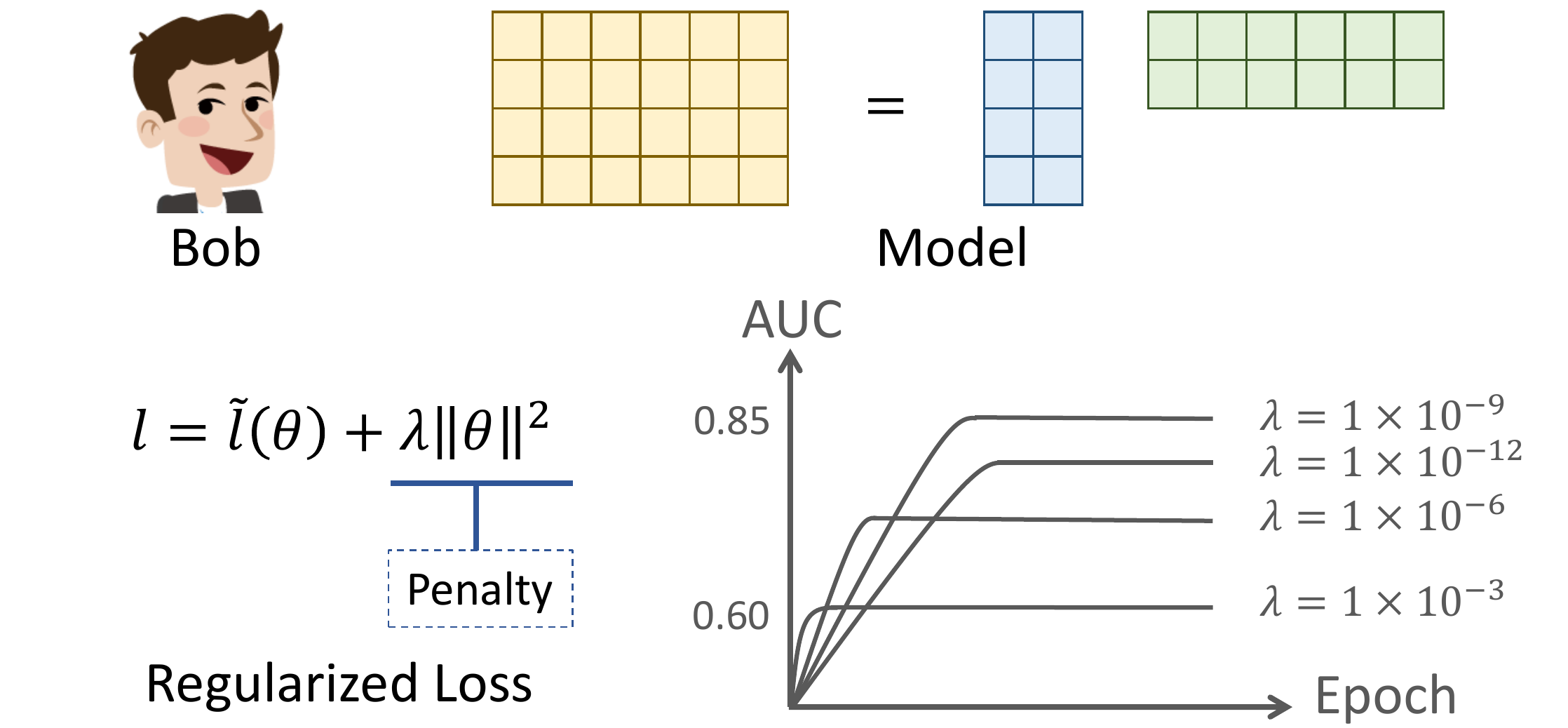}
    \vspace{-2mm}
    \caption{A motivating example of regularization tuning: the model is highly sensitive to the choice of $\lambda$.} \vspace{-6pt}
    \label{fig:bob}
\end{figure}

Recommender systems typically work with a large number of categorical variables, such as user/item ID, user demographics, and item tags. 
Conventionally, these categorical variables are handled by the techniques of one-hot encoding or embedding. One-hot encoding converts a categorical variable into a set of binary variables, while the embedding technique projects each categorical value into a latent vector space. 
Since some categorical variables might have high cardinality (like ID features), there may not be sufficient data to learn the feature interactions. As such, recommender models trained with either technique could be prone to overfitting~\cite{he2017neural}. Moreover, the interactions among the categorical features are usually long-tailed \cite{dyer2011visualizing}, with a head made up of highly frequent values and a long tail of rare ones.
For example, consider the interaction between user ID and the buy-or-not variable, most (user ID, buy-or-not) pairs appear less than 50 times and there are very few pairs have more than 100 occurrences.
The data sparsity issue caused by high-cardinality features and non-uniform occurrences pose practical challenges to train recommender models, making appropriate regularization essential~\cite{tutorial2011}. In fact, the performance of many recommender models, even the simple matrix factorization model, varies widely depending on the regularization setting. 
As a result, manually tuning the regularization hyperparameters could be extremely hard for practitioners with little experience, and even non-trivial for experienced researchers in industry. Methods like grid search might help but at the inevitable cost of large computation resources. Figure \ref{fig:bob} shows a motivating example. \vspace{+5pt}

\noindent\textbf{Motivating Example}. \noindent\emph{After rounds of interviews, Bob finally gets a job as a machine learning engineer specialized on news recommendation. This week, he plans to experiment the matrix factorization model, which is reported to outperform the production model in literature. However, as experiments go on, he finds the models are \textbf{large} on the company data, with millions of parameters. "Hmm, I need some regularization!", he thinks. He then adds $L_2$ regularizer on the embedding parameters, but wondering how to choose the regularization coefficient $\lambda$? "Whatever it is, let me have a try first!", Bob then \textbf{randomly} chooses some values of $\lambda$. To his surprise, different choices of $\lambda$ lead to more than 30\% fluctuation in model performance.}

\noindent\emph{The observation of \textbf{high sensitivity} to $\lambda$ motivates Bob to search for more \textbf{fine-grained} $\lambda$, as he believes that good $\lambda$ can substantially boost the performance. 
Instead of applying a uniform $\lambda$ on all embedding parameters, he considers varying it by the embedding dimensions. 
However, he has other tasks unfinished so he can't babysit the tuning process. He decides to employ grid search with 10 candidate values. 
As the embedding size $K=128$, dimension-wise $\lambda$ would take about $10^{128}$ \textbf{full training runs}! So all Bob can do now is to buy new machines or babysit the tuning process by himself...}\vspace{+5pt}

Examples similar to the above are not rare in industry. Practitioners like Bob, who often expect a high salary and a fantastic job with great intellectual challenges, would find their expectations clashing with the reality, spending most of the time on the tedious job of hyperparameter tuning. When it comes to recommender systems, tuning the regularization has been a nightmare for many practitioners whenever a new model is to be launched. 
Despite the high value of regularization tuning, there is relatively little research to conquer this issue. Methods like grid search could alleviate the laborious tuning process, but at a very high cost, especially when we want to tune $\lambda$ in a finer granularity. 
Other automated hyperparameter selection methods are also computation-expensive, since they typically require multiple full training runs \cite{shahriari2016taking,snoek2012practical}. {An automatic method that can find appropriate $\lambda$ on the fly with affordable cost would be more than a blessing to practitioners.} 

Prior work on automatic regularization on recommender models is scarce. The most relevant method is SGDA\cite{Rendle2012LearningRegularization}, which is a dimension-wise adaptive regularization method based on stochastic gradient descent (SGD). However, SGDA is designed for the task of rating prediction rather than personalized ranking, applying it would result in weak top-N recommendation performance~\cite{Cremonesi:2010}.  
In addition, we argue that more fine-grained $\lambda$, such as user-wise, is more advantageous than dimension-wise since it can adapt the regularization strength for users of different activity levels. 
Furthermore, adaptive optimizers like Adam~\cite{kingma2014adam} and Adagrad~\cite{duchi2011adaptive}, are more effective and converge faster than SGD in optimizing recommender models~\cite{NFM}. 
As such, we believe that there is an urgent need to develop an adaptive regularization method for top-K recommendation, and more importantly, should support fine-grained tuning with adaptive optimizers, instead of being merely applicable to dimension-wise tuning and plain SGD.

\noindent\textbf{Contributions}.
In this paper, we explore how to design an automatic method to regularize recommender models. 
Focusing on the personalized ranking task~\cite{RendleBPR:Feedback}, 
we propose \textsc{$\lambda$Opt}, a generic regularizer that learns the regularization coefficients during model training based on validation data. 
The basic idea is to employ Bayesian Personalized Ranking (BPR) loss~\cite{RendleBPR:Feedback} on validation data as the objective function, treating the regularization coefficients as the variable to the function and optimizing it with gradient descent. 
We illustrate our approach on matrix factorization, which is representative of a large family of embedding-based recommender models~\cite{bayer2017generic}. 
We highlight
the elements that distinguish \textsc{$\lambda$Opt} as follows: 
\begin{itemize}[leftmargin=*]
    \item  \textsc{$\lambda$Opt} adaptively finds the appropriate $\lambda$. It enjoys substantially lower computation cost compared to other automated methods~\cite{shahriari2016taking,snoek2012practical} that require multiple training runs. 
    \item By virtue of automatic differentiation, \textsc{$\lambda$Opt} obviates the complex derivations of gradients. Hence it can be conveniently generalized to a diverse set of recommender models.
    \item By permitting advanced optimizers with adaptive learning rates, \textsc{$\lambda$Opt} overcomes the issue of limited optimizer choice in \cite{Rendle2012LearningRegularization}, making it more practitioner-friendly.
    \item Last but not least, our design of \textsc{$\lambda$Opt} facilitates regularization in any granularity. $\lambda$ can be dimension-wise, user-wise, item-wise or any combinations among them. Such fine-grained regularization brings considerable benefits to recommendation performance.
\end{itemize}

We conduct extensive experiments on \textsc{$\lambda$Opt} to justify its effectiveness in regularizing recommender models. We find that models trained with \textsc{$\lambda$Opt} significantly outperform grid search (fixed $\lambda$) and SGDA~  \cite{Rendle2012LearningRegularization}, demonstrating high utility of \textsc{$\lambda$Opt}. 
To sum up, \textsc{$\lambda$Opt} is a simple yet effective training tool, which not only lowers the barrier for practitioners to launch their recommender models but also boosts the performance with fine-grained regularization.

\section{Preliminaries}

\begin{figure*}
\centering
\vspace{-2mm}
\subfigure[Fixed approach]{\includegraphics[width=0.36\linewidth]{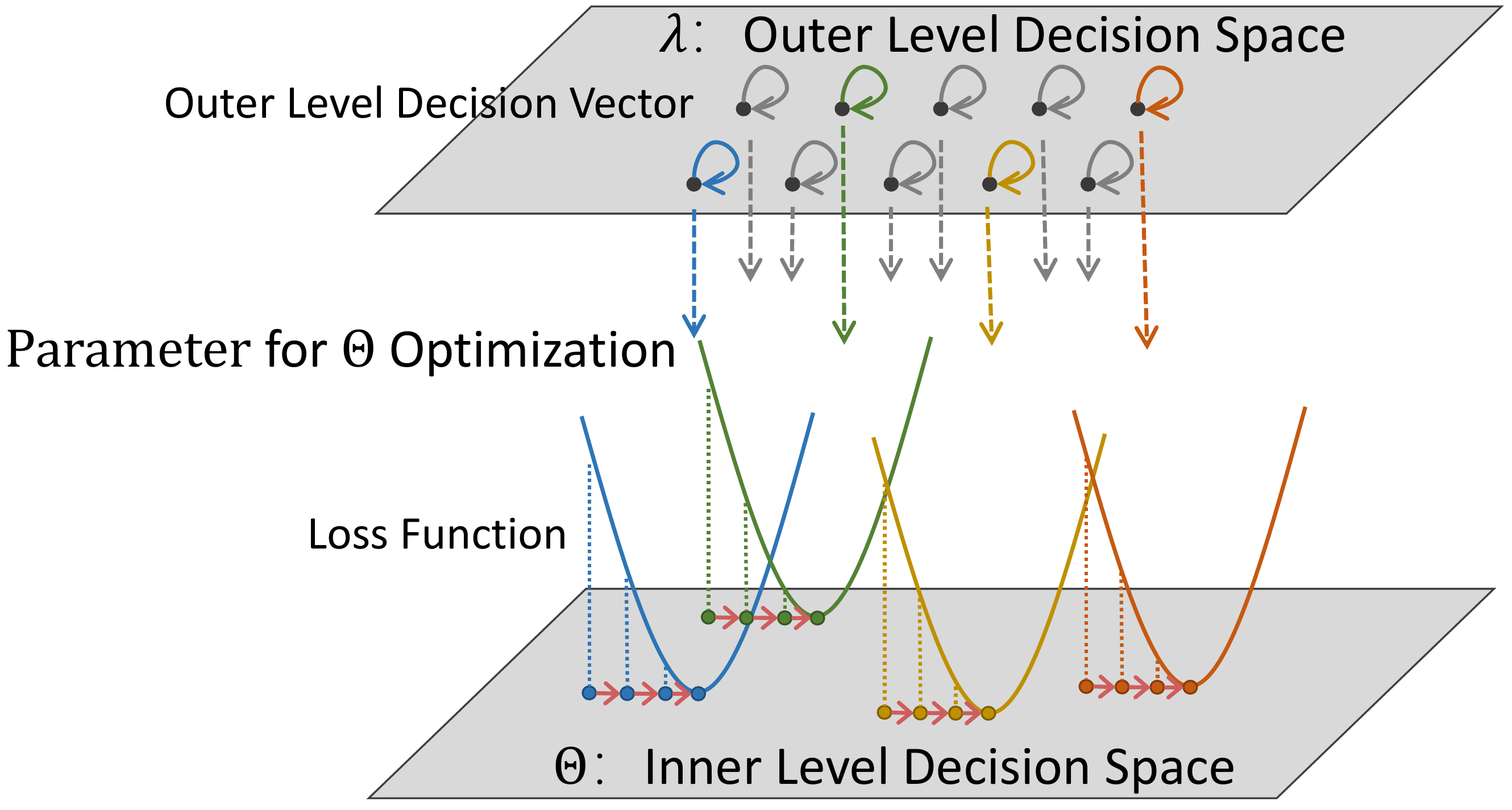}}
\hspace{6mm}
\subfigure[Adaptive approach]{\includegraphics[width=0.36\linewidth]{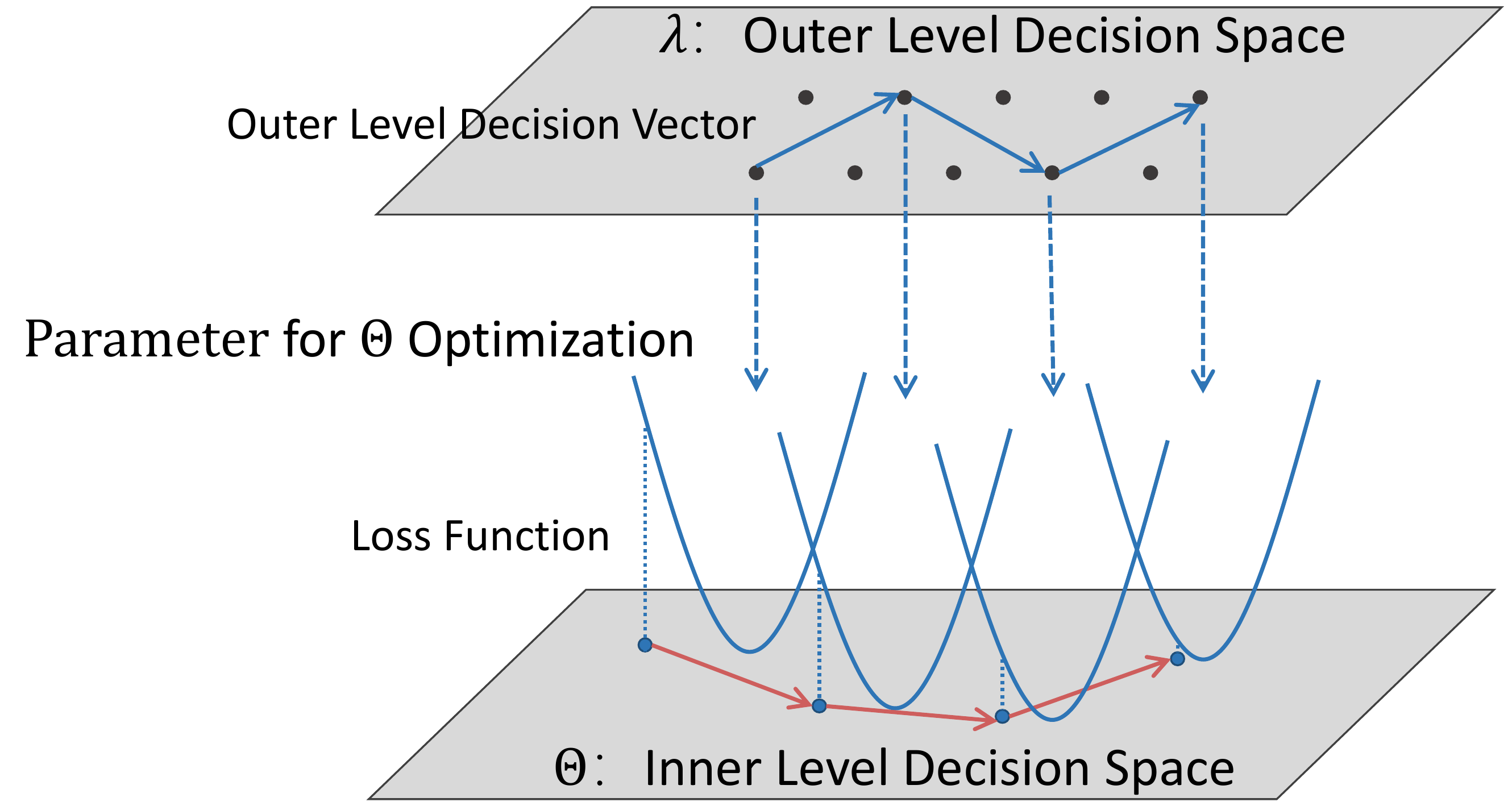}}
\vspace{-4mm}
\caption{$\Lambda$-trajectory for fixed regularization approach and adaptive regularization approach.}
\vspace{-2mm}
\label{fig:process}
\end{figure*}

\subsection{Matrix Factorization}
Matrix Factorization (MF) \cite{koren2009matrix} plays a dominant role in recommender systems. The basic principle behind MF is that we could project users or items into a latent space so that users' preferences can be reflected using their proximity to the items. 
In spite of their prevalence, previous work on such matrix factorization models observes that {the choice of the regularization coefficients $\lambda$ has significant influence on the trained models.} 
Actually, this phenomenon stems from the inherent structure of matrix factorization models, --- the number of parameters is often much larger than the number of samples, and the characteristics of recommendation datasets, -- activity levels of users/items can be extremely diverse. Small $\lambda$ leads to overfitting while large $\lambda$ might cause underfitting. For this reason, it is no surprising that the problem of tuning regularization coefficients has been of extreme importance in practice.

\subsection{Bayesian Personalized Ranking}
Top-K item recommendation from implicit feedback is a prevalent task in real-world recommender systems~\cite{chen2017attentive,yu2018aesthetic,zhang2017joint}. 
With Bayesian Personalized Ranking (BPR) \cite{RendleBPR:Feedback} as the optimization objective, we study the adaptive regularization for factorization models. 
Targeting at learning from implicit feedback, 
BPR assumes that the user $u$ prefers the observed items over all the other unobserved ones. Formally, it aims to minimize the objective function: 
\begin{align}
l_{S_T}(\Theta|\lambda) &= \tilde{l}_{S_T}(\Theta) + \Omega(\Theta|\lambda) \\
&= - \sum_{(u,i,j) \in S_T} \ln (\sigma(\hat{y}_{ui}(\Theta) - \hat{y}_{uj}(\Theta))) + \Omega(\Theta|\lambda),
\label{eq:bpr loss}
\end{align}
where $\Theta$ denotes the model parameters, $\lambda$ denotes the regularization coefficients, and $\sigma(\cdot)$ denotes the sigmoid function. $S_T=\{(u,i,j)|i\in \mathcal{I}_u \wedge j\in \mathcal{I} \backslash \mathcal{I}_u\}$ denotes the dataset of training triplets, where $\mathcal{I}_u$ represents the set of the observed items that user $u$ has interacted with in the past, and $\mathcal{I}$ denotes the set of all items. $\hat{y}_{ui}$ scores $(u,i)$ and can be parameterized using factorization models. $\Omega(\Theta|\lambda)$ is the penalty term indicating regularization on the model. Usually, the objective function is optimized by stochastic gradient descent (SGD). In addition to standard SGD optimizer, advanced optimizers that adapt their learning rate during training, for example, Adam \cite{kingma2014adam}, can be used to accelerate the training process.

\section{Methodology}

In conventional regularization tuning, the goal is to find the optimal $\Lambda$ to train a regularized model, which achieves best performance on the validation set\footnote{As we focus on fine-grained regularization, in the following paper, $\Lambda$ and $\lambda$ can be either a vector or matrix. }. Essentially, this can be formulated as a nested optimization problem \cite{RendleBPR:Feedback} or bi-level optimization problem \cite{sinha2018review}:
\begin{equation}
\begin{split}
\min_{\Lambda} \sum_{(u', i', j') \in S_V} l(u', i', j' |\arg \min_{\Theta} \sum_{(u, i, j) \in S_T}l(u, i, j|\Theta, \Lambda)), 
\end{split}
\label{eq:nested}
\end{equation}
where the inner level attempts to minimize training loss with respect to $\Theta$ while the outer level addresses the minimization of validation loss with respect to $\Lambda$ on validation set $S_V$. A naive solution would be exhaustive search, training a regularized model for every possible $\Lambda$ and then choosing the one with the best validation performance. In practice, due to time and resources constraints, people resort to methods like grid-search, sampling $\Lambda$ from a small but reasonable interval. However, as model sizes and dataset volumes increase, grid search might also be unaffordable, particularly for large-scale applications, such as real-world recommendation.

In order to efficientize the search, previous work proposed to alternate the optimization of $\Lambda$ and $\Theta$, between consecutive full training runs \cite{chapelle2002choosing, larsen1998adaptive}, or on the fly \cite{Rendle2012LearningRegularization, Luketina2016ScalableHyperparameters}. Compared to grid-search, where $\Lambda$ is fixed during a full training run, the on-the-fly adaptive methods in \cite{Rendle2012LearningRegularization, Luketina2016ScalableHyperparameters} adjusts $\Lambda$ according to performance on validation sets every training step. Nevertheless, we show here that both approaches can be interpreted as attempting to find a plausible trajectory in the $\Lambda$ space to regularize the model well, and we introduce our proposed method later.

\subsection{$\Lambda$-Trajectory}
\begin{definition}
 A $\Lambda$-trajectory is the sequence of regularization coefficients for a full training run: $\Lambda = \{\Lambda_1, ... \Lambda_T\}$,
where $T$ denotes the total steps in the training run.
\end{definition}
Figure \ref{fig:process} visualizes $\Lambda$-trajectories for fixed approach and adaptive approach. In the nested optimization problem, there is an outer level decision space ($\Lambda$ space in our case) and an inner level decision space ($\Theta$ space in our case) \cite{sinha2018review}. Once $\Lambda$ is selected, it becomes a fixed parameter in the loss function for $\Theta$ optimization, indicated by the curve between the inner level decision space and outer level decision space in the figure. $\Lambda$-trajectories are represented using arrow lines in $\Lambda$ space with arrows indicating the update directions of $\Lambda$ at each training step. 

\subsubsection{Fixed $\Lambda$ Approach}
In fixed $\Lambda$ approach, such as grid-search, we have the same regularization coefficients for all training steps, which are manually set at the beginning of training: $\Lambda_1 = \Lambda_2 = ... = \Lambda_T$. 
Generally, grid-search goes as follows. We first select a few candidates from $\Lambda$ space. Then, we accordingly train multiple models and compare their performances on the validation set to choose the best $\Lambda$. The process can be regarded as trying out multiple $\Lambda$-trajectories while all of them are restricted -- going around in circles as indicated in Figure \ref{fig:process}(a). Due to the constraint on each $\Lambda$-trajectory, we usually need to search over a good number of $\Lambda$ candidates before $\Lambda$ space is explored sufficiently and appropriate regularization is achieved. The search either takes expensive computation or requires prior knowledge about picking suitable $\Lambda$ candidates. Even worse, if we consider regularization in finer levels, the computation of searching over all the selected $\Lambda$ would make grid-search barely feasible.

\subsubsection{Adaptive $\Lambda$ Approach}
As shown in Figure \ref{fig:process}(b), the adaptive approach in \cite{Rendle2012LearningRegularization, Luketina2016ScalableHyperparameters}, instead, employs different regularization coefficients at each training step, allowing faster exploration in $\Lambda$ space. We justify intuitions behind the adaptive approach here, which motivate us to adopt the adaptive paradigm when we design \textsc{$\lambda$Opt}.
\begin{itemize}[leftmargin=*]
    \item At different training stages, the strength of regularization should be different. For instance, there should be little regularization at the early stages of training since the model has not learned much from the data while strong regularization might be necessary for the late stage after the model sees the data a great many times.
    \item Assuming we have sufficient validation data, adjusting $\Lambda$ based on the validation performance would cause no obvious overfitting to the validation set.
    \item Although the resulted $\Lambda$-trajectory might not be optimal, it would be a competent one with substantially lower computation.
\end{itemize}  

\begin{figure}
    \centering
    \includegraphics[width=0.475\textwidth]{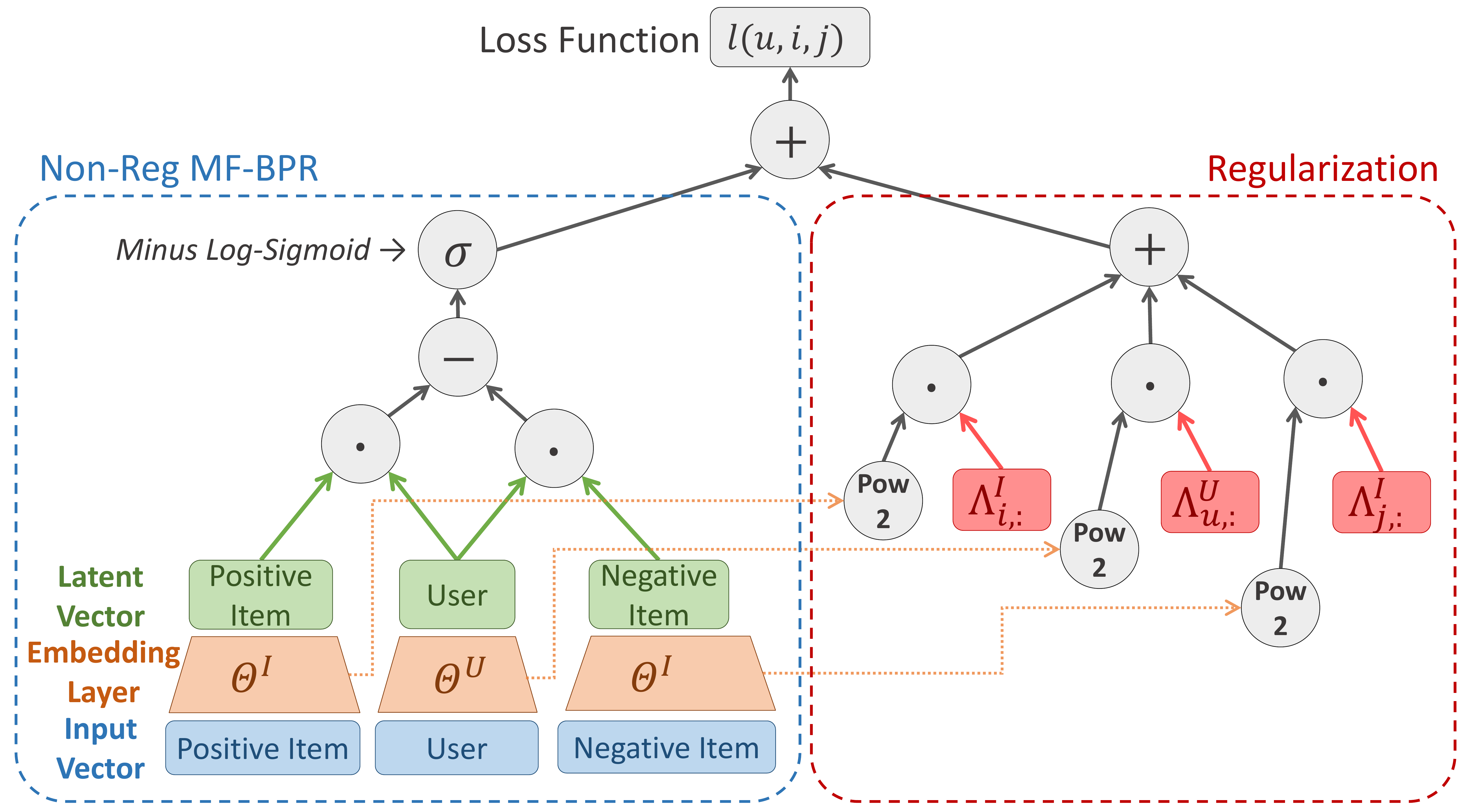}
    \vspace{-6mm}
    \caption{MF-BPR with Fine-grained Regularization.}
    \label{fig:model}
    \vspace{-3mm}
\end{figure}

\subsection{\textsc{$\lambda$Opt}: Efficient Exploration over $\Lambda$ Space}
In order to generate $\Lambda$-trajectories that steer efficient exploration over $\Lambda$ space, \textsc{$\lambda$Opt} takes into account the two aspects -- $\Lambda$ in finer levels and its adaptive update. As we have just explained in Section 3.1, adaptive update of $\Lambda$ accelerates the exploration over $\Lambda$ space due to its flexibility in changing $\Lambda$ in a single training run. In this subsection, we show how fine-grained $\Lambda$ contributes to the exploration and present \textsc{$\lambda$Opt} formally. 

\subsubsection{Recommender Models with Fine-grained Regularization}

As shown in Figure \ref{fig:model}, traditional matrix factorization with BPR (MF-BPR) consists of a non-regularized block and a regularized block. Distinguished from common regularization strategies, setting a global $\lambda$ or dimension-wise $\lambda$, we consider regularization in finer levels -- the regularized part in our method is user-/item-aware instead of being only dimension-wise. Given the fact that recommender systems interact with heterogeneous users and items, incorporating user-/item-aware regularization into \textsc{$\lambda$Opt} can be vital. Besides, from the perspective of $\Lambda$-trajectories, entailing the same regularization for each user/item, dimension-wise $\lambda$ or global $\lambda$ equivalently forbids exploration in directions related to users/items, thus preventing the discovery of better regularization strategies to train recommender models.  

\subsubsection{Regularizer Endowed with Adaptive Regularization in Finer Levels} In this subsection, we show how we derive the adaptive update of $\Lambda$ in our method. 
We choose to implement $\textsc{$\lambda$Opt}$ as neural networks with $\Lambda$ as weights. Such design is more practitioner-friendly compared to SGDA, since it saves them from complex derivations of the gradients and makes $\textsc{$\lambda$Opt}$ easy to generalize across various models, loss functions and model optimizers. 

The nested optimization problem in Equation \ref{eq:nested} is hard to solve directly. Alternating optimization \cite{Rendle2012LearningRegularization} reduces it into two simpler ones. To be exact, we perform the following two steps iteratively
\begin{itemize}[leftmargin=*]
    \item \textbf{$\Theta$ Update at Step t}. We fix $\Lambda_t$ while $\Theta$ is optimized using triplets $(u, i, j)$ sampled from train set $S_T$. Notice that we employ $\Lambda$ in finer levels to compute the regularized training loss.
    \item \textbf{$\Lambda$ Update at Step t}. We fix $\Theta_t$ while $\Lambda$ is computed by optimizing the expected validation loss with triplets $(u, i, j)$ sampled from validation set $S_V$.
\end{itemize}

Since our goal is to find $\Lambda$ which would achieve the smallest validation loss, we need to figure out the relationship between $\Lambda$ and the validation loss. As pointed out in \cite{Rendle2012LearningRegularization}, while the validation loss for current model $l_{S_V}(\Theta_{t})$ has nothing to do with $\Lambda_t$, the expected validation loss $l_{S_V}(\Theta_{t+1})$ instead depends on $\Lambda_t$ if $\Theta_{t+1}$ is obtained using $\Lambda_{t}$. This suggests that we can first obtain the assumed next-step model parameters $\bar{\Theta}_{t+1}$ and then compute the validation loss with $\bar{\Theta}_{t+1}$ and its gradients with respect to $\Lambda$. We use the word "assumed" because this update is never really performed on the model we finally want. We only use it to obtain the direction to update $\Lambda$ i.e.  to move a step in $\Lambda$-trajectory. We use symbols in the format of $\bar{\cdot}$ to distinguish "assumed" ones from ordinary ones.

\begin{figure*}
    \centering
    \includegraphics[width=0.98\textwidth]{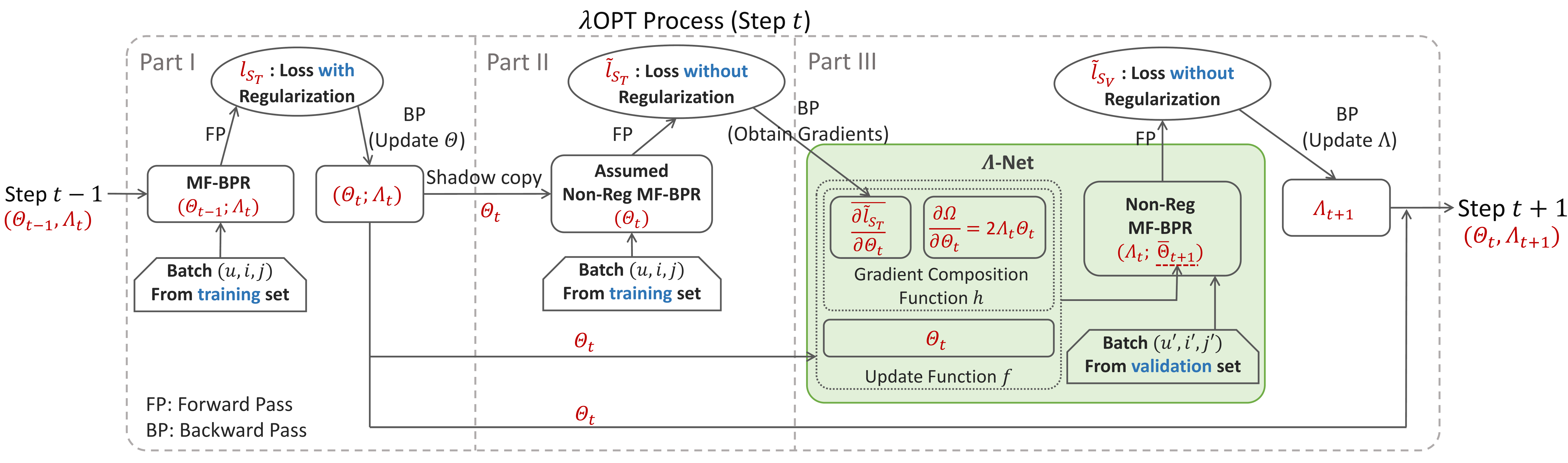}
    \caption{Regularizer Endowed with Adaptive Regularization in Finer Levels.}
    \label{fig:regularizer}
\end{figure*}

\subsubsection*{\textbf{Obtain Assumed Next-Step Model Parameters $\bar{\Theta}_{t+1}$.}} The key to obtain $\bar{\Theta}_{t+1}$ is to compute the gradients of assumed regularized training loss with respect to $\Theta_t$. \textsc{$\lambda$Opt} tackles this via splitting the gradients into two terms, one for non-regularized loss and the other for the penalty term:
\begin{equation}
\overline{\frac{\partial{l_{S_T}}}{\partial\Theta_t}} = h(\overline{\frac{\partial \tilde{l}_{S_T}}{\partial\Theta_t}}, \Theta_t) = \overline{\frac{\partial \tilde{l}_{S_T}}{\partial\Theta_t}} + \frac{\partial \Omega}{\Theta_t}.
\label{eq:gradient sum}
\end{equation}
We denote as $h$ the function composing non-regularized gradients and regularized gradients. In fine-grained $L_2$ regularization, $\Omega(\Theta|\Lambda) = \Lambda ||\Theta||_2^2$, Equation \ref{eq:gradient sum} would be 
\begin{equation}
\overline{\frac{\partial{l_{S_T}}}{\partial\Theta_t}} = h(\overline{\frac{\partial \tilde{l}_{S_T}}{\partial\Theta_t}}, \Theta_t) = \overline{\frac{\partial \tilde{l}_{S_T}}{\partial\Theta_t}}  + 2\Lambda\Theta_t.
\label{eq:gradient sum l2}
\end{equation}
As we don't want assumed gradients to mess around with the ones in MF-BPR, \textsc{$\lambda$Opt} performs a separate forward \& backward computation on $S_T$ to obtain $\overline{\frac{\partial \tilde{l}_{S_T}}{\partial\Theta_t}}$ as illustrated in Part II of Figure \ref{fig:regularizer}. After composition of assumed regularized gradients using $h$, the assumed next-step model parameters are given by $\bar{\Theta}_{t+1} = f(\Theta_t, \overline{\frac{\partial l_{S_T}}{\partial\Theta_t}})$ where $f$ is the update function of $\Theta$, generally determined by the optimizer $\Theta$ update used.

\subsubsection*{\textbf{Minimize the Validation Loss}} Up to now, we have obtained assumed next-step model parameters $\bar{\Theta}_{t+1}$ and the only remaining job is to find the $\Lambda$ which minimizes validation loss $l_{S_V}(\bar{\Theta}_{t+1})$. Note that this is a constrained minimization though treated as unconstrained in SGDA \cite{Rendle2012LearningRegularization}. Mathematically, we want to solve
\begin{equation}
\begin{split}
\arg\min_{\Lambda} - \sum_{(u, i, j) \in S_V} \ln (\sigma(\hat{y}_{ui}(\bar{\Theta}_{t+1}) - \hat{y}_{uj}(\bar{\Theta}_{t+1})),\\
\text{subject to } \Lambda \geq 0.\\
\end{split}
\label{lambda update objective}
\end{equation}
Karush-Kuhn-Tucker (KKT) conditions for constrained minimization with non-convex objectives give feasible regions in $\Lambda$ space, which make the search more efficient and stable. The gradients of validation loss with respect to $\Lambda$ are denoted as $G$
\begin{equation}
G=\nabla_{\Lambda} - \sum_{(u, i, j)\in S_V}\ln (\sigma(\hat{y}_{ui}(\bar{\Theta}_{t+1}) - \hat{y}_{uj}(\bar{\Theta}_{t+1})).
\end{equation}
Then KKT gives 
\begin{equation}
\Lambda G = 0, G \ge 0, \Lambda \ge 0. 
\label{feasible lambda}
\end{equation}
From Equations \ref{feasible lambda}, we can see that feasible solutions of $\Lambda$ require both $G$ and $\Lambda$ to be non-negative with one of them equals to zero. A slack version could be encouraging $G$ to be small and $\Lambda$ to be non-negative. 

Part III of Figure \ref{fig:regularizer} shows the computation flow of minimizing validation loss in \textsc{$\lambda$Opt}. We term the entire block in green colour as $\Lambda$-Net because the parameters here are $\Lambda$ rather than $\Theta$. After forward \& backward passes over $\Lambda$-Net, we add the slack constraints given by Equation \ref{feasible lambda}: clip the gradients $G$ to a small value and smooth the negative entries in $\Lambda$ as zero. The updated parameters of $\Lambda$-Net serve as the regularization coefficients $\Lambda_{t+1}$ for MF-BPR in next iteration.

\subsubsection*{\textbf{Computation Cost}} 
$\textsc{$\lambda$Opt}$ adjusts $\Lambda$ on the fly, which obviates multiple full training runs as grid-search like methods. Since one iteration takes 3 forward passes and backpropagations, the computation cost for a single training run is only 3 times the one for the fixed approach. For practitioners, this is cost-effective because they do not have to search over a great many of $\Lambda$ candidates. 

\section{Applications}
In this section, we show how to apply \textsc{$\lambda$Opt} with various model optimizers and fine-grained regularization. We start from the simple matrix factorization model. Generally, the model parameters, user embedding $\Theta^U$ and item embedding $\Theta^I$, should be matrices of $|U| \times K$ and $|I| \times K$ respectively, where $K$ stands for number of dimensions. As stated above, we only need to specify the forward pass for the model and optimizer update. For MF, the forward pass to compute non-reg BPR loss is 
\begin{equation}
\tilde{l}(u, i, j, \Theta|\Lambda) = - ln(\sigma(\Theta^U_u \cdot \Theta^I_i - \Theta^U_u \cdot \Theta^I_j)).
\end{equation}
The model parameters are $\Theta = \{\Theta^U, \Theta^I\}$. Correspondingly, the regularization coefficients are $\Lambda = \{\Lambda^U, \Lambda^I\}$.

\subsection{Optimizer Choices}
When training recommender models, various optimizers can be used. Model parameters update of these optimizers are often different. Unlike SGDA \cite{Rendle2012LearningRegularization} and \cite{Luketina2016ScalableHyperparameters}, \textsc{$\lambda$Opt} can cope with various optimizers as long as the model parameter update function $f$ is derivable with respect to $\Lambda$. For example, if we use a SGD optimizer, 
\begin{equation}
\bar{\Theta}_{t+1} = f(\Theta_t, \overline{\frac{\partial l_{S_T}}{\partial \Theta_t}}) = \Theta_t - \eta h(\overline{\frac{\partial {l}_{S_T}}{\partial \Theta_t}}, \Theta_t).
\end{equation}
Substituting Equation \ref{eq:gradient sum} and Equation \ref{eq:gradient sum l2} into the above equation, we can obtain
\begin{align}
\bar{\Theta}_{t+1} & = f(\Theta_t, \overline{\frac{\partial l_{S_T}}{\partial \Theta_t}}) = \Theta_t - \eta \overline{\frac{\partial \tilde{l}_{S_T}}{\partial \Theta_t}} - 2\eta\Lambda\Theta_t.
\end{align}
In order to obtain the validation loss's gradient with respect to $\Lambda$, we need to compute $\frac{\partial f}{\partial \Lambda}$. For the simple SGD optimizer, this would be easy: $-2\eta\Theta_t$. However, it is not the case for complex optimizer like Adam, which is among the off-the-shelf choices when practitioners start to train their factorization models --- they would have to derive the gradients themselves! Luckily, in \textsc{$\lambda$Opt}, this step of obtaining gradients would be handled by Automatic Differentiation framework such as TensorFlow\footnote{https://www.tensorflow.org/} \cite{abadi2016tensorflow} and PyTorch\footnote{https://pytorch.org/}. Hence, we don't need to worried about the complex derivation of $\frac{\partial f}{\partial \Lambda}$. For Adam optimizer, we only need to specify $f$ as follows:
\begin{align}
 \bar{\Theta}_{t+1} &= \Theta_t - \eta \frac{\sqrt{1 - \beta_2^t}}{\sqrt{1 - \beta_1^t}}\frac{s_t}{\sqrt{r_t} + \epsilon}, \\
 s_t &= \beta_1 s_{t-1} + (1 - \beta_1)\overline{\frac{\partial l_{S_T}}{\partial \Theta_t}}, \\
 r_t &= \beta_1 r_{t-1} + (1 - \beta_1)\overline{\frac{\partial {l_{S_T}}}{\partial \Theta_t}} \odot \overline{\frac{\partial {l_{S_T}}}{\partial \Theta_t}}.
\end{align}
\subsection{Fine-grained Regularization}
To our knowledge, no previous work has explored regularization in finer levels, like user-/item-aware regularization. Grid-search fails due to unaffordable computation cost of searching over extremely high-dimensional $\Lambda$ spaces. And SGDA is also not applicable since its derivation only considers dimension-aware regularization but does not adapt with users and items. As stated in Section 3.2.1, our design of \textsc{$\lambda$Opt} naturally lends itself to user/item-aware $\Lambda$. Fine-grained regularization, for example user-wise regularization, could be done by expanding the penalty term and obtaining the gradients as following:
\begin{align}
\Omega(\Theta^U, \Theta^I|\Lambda^U, \lambda^I) &=  \sum_{u=1}^{|U|} \Lambda^U_{u} \sum_{k=1}^K(\Theta^U_{u,k} )^2 + \lambda^I ||\Theta^I||_2^2, \\
\frac{\partial \Omega}{\partial \Theta^U_{u,k}}& = 2\Lambda_u^U\Theta_{u,k}^U, 
\frac{\partial \Omega}{\partial \Theta^I} = 2 \lambda^I \Theta^I,
\end{align}
where $\Lambda^U$ is a $|U| \times 1$ vector specifying the user-wise regularization for the user embedding matrix and $\Lambda^I$ is a scalar specifying the regularization for item embedding matrix. Similarly, we can use more fine-grained regularization $\Lambda$ that combines dimension-wise, user-wise and item-wise.
\begin{gather}
\!\!\!\!\!\Omega(\Theta^U, \Theta^I|\Lambda^U, \Lambda^I)\! =\!\! \sum_{u=1}^{|U|}\!\sum_{k=1}^K \!\Lambda^U_{u,k}(\Theta_{u,k}^U)^2 
\!+\!
\sum_{i=1}^{|I|}\!\sum_{k=1}^K \Lambda^I_{i,k}(\Theta_{i,k}^I)^2, \\
\frac{\partial \Omega}{\partial \Theta^U_{u,k}} = 2\Lambda_{u,k}^U\Theta_{u,k}^U,~~~~~ 
\frac{\partial \Omega}{\partial \Theta^I_{i,k}} = 2\Lambda_{i,k}^I\Theta_{i,k}^I,
\end{gather}
where $\Lambda^U$ and $\Lambda^I$ are $|U| \times K$ and $|I| \times K$ matrices respectively.

\section{Empirical Study}
In this section, we empirically evaluate our methods with the aim of answering the following research questions: 
\begin{itemize}
\item[RQ.1] What is the performance of MF models trained using \textsc{$\lambda$Opt}?  The adaptive method can save practitioners a lot time. But does it come at a cost of worse performance compared to other regularization strategies?
\item[RQ.2] With \textsc{$\lambda$Opt}, practitioners can add fine-grained regularization over MF models conveniently, which is infeasible in grid-search like methods. Will such fine-grained regularization be effective in addressing heterogeneous users and items? 
\item[RQ.3] What are the $\Lambda$ trajectories of \textsc{$\lambda$Opt} like? Does \textsc{$\lambda$Opt} find better trajectories to regularize the model? The $\Lambda$ trajectories can explain the performance difference across users and items with varied frequency, telling us why \textsc{$\lambda$Opt} performs better or worse than the fixed approaches.
\end{itemize}

\subsection{Experimental Settings}
\subsubsection{Datasets}
We experiment with two public datasets: Amazon Food and MovieLens 10M. Table \ref{tab:data summary} summarizes the statistics of datasets after pre-processing.

- \textbf{Amazon Food Review}\footnote{https://www.kaggle.com/snap/amazon-fine-food-reviews}. It contains reviews of fine foods from amazon, spanning a period of more than 10 years. We filter the dataset and only keep the users and items with more than 20 records. We omit the exact scores and treat every entry in the dataset as a positive sample. 

- \textbf{MovieLens 10M}\footnote{https://grouplens.org/datasets/movielens/10m/}. It is a widely used benchmark dataset and contains timestamped user-movie ratings ranging from $1$ to $5$. We use the ``10M'' version, which contains approximately $10$ million ratings drawn from 69,878 users. Each user has at least 20 ratings. We use it as an implicit feedback dataset, where the exact ratings are omitted and each entry is regarded as a positive sample. 

\begin{table}[tbp]
\centering
\caption{Statistics of datasets.}
\vspace{-1mm}
\label{tab:data summary}
{\scalebox{1}{
\begin{tabular}{c|cccc}
\hline
Dataset & \# User & \# Item & \# Interaction &  Density \\
\hline
Amazon Food & $1, 238$ & $3,806$ & $38,919$ & $0.825\%$\\
MovieLens 10M & $69,878$ & $10,677$ & $10,000,054$ & $1.340\%$ \\
\hline
\end{tabular}}}
\vspace{-1mm}
\end{table}

\subsubsection{Performance Measures}
For both Amazon Food Review and MovieLens 10M, we divide the data according to the time stamp information. Specially, for each user, all the records are divided into training, validation and testing set based on the proportion $60\%$, $20\%$ and $20\%$. To evaluate the performance of our methods, for each (user, item) pair in the test set, we make recommendations by ranking all the items that are not interacted by the user in the training and validation set. Three metrics are evaluated:

- \textbf{AUC}. Area under the Receiver Operating Characteristic or ROC curve (AUC) means the probability to rank a randomly chosen positive item higher than a randomly chosen negative item. 

- \textbf{HR}. Hit Ratio (HR) is based on recall. It intuitively measures whether the test item is in the top-K list. We set K, the truncation length of the ranking list, to be $50$ (HR@50) and $100$ (HR@100).

- \textbf{NDCG}. Normalized Discounted Cumulative Gain (NDCG) considers positions of hits in the top-K list, where hits at higher positions get higher scores. K is as stated above.

For HR and NDCG, we report the average score of all the users. The score for each user is averaged over all his/her test items. For all the metrics, higher score is better.

\begin{table*}[tbp]
\centering
\caption{Recommendation Performance on Amazon Food Review and MovieLens 10M.}
\label{tab:results}
{\scalebox{1}{
\begin{tabular}{ll|ccccc|ccccc}
\hline
\multicolumn{2}{c|}{\multirow{2}{*}{Method}}  & \multicolumn{5}{c|}{Amazon Food Review}& \multicolumn{5}{c}{MovieLens 10M}\\ \cline{3-12}
 & & AUC & \!HR@50 & \!HR@100 & \!NDCG@50 & \!NDCG@100 & AUC & \!HR@50 & \!HR@100 & \!NDCG@50 & \!NDCG@100 \\
 \hline
\multicolumn{2}{l|}{\!SGDA \cite{Rendle2012LearningRegularization}} & 0.8130 & 0.1786 & 0.3857 & 0.1002 & 0.1413 & 0.9497 & 0.2401 & 0.3706 & 0.0715 & 0.0934\\
\multicolumn{2}{l|}{\!AMF \cite{he2018adversarial}  }& 0.8197 & 0.3541 & 0.4200 & 0.2646 & 0.2552 & 0.9495 & 0.2625 & 0.3847 & 0.0787 & 0.0985\\
\multicolumn{2}{l|}{\!NeuMF \cite{he2017neural} }& 0.8103 & 0.3537 & 0.4127 & 0.2481 & 0.2218 & 0.9435 & 0.2524 & 0.3507 & 0.0760 & 0.0865\\
\!MF-$\lambda$\textsc{Fix} && 0.8052 & 0.3482 & 0.4163 & 0.2251 & 0.2217 & 0.9497 & 0.2487 & 0.3779 & 0.0727 & 0.0943\\
\hline
\!MF-$\lambda$\textsc{Opt} &\!\!\!-D\!& 0.8109 & 0.2134 & 0.3910 & 0.1292 & 0.1543 & 0.9501& 0.2365 & 0.3556 & 0.0715 & 0.0909\\
&\!\!-DU& 0.8200 & 0.3694 & 0.4814 & 0.2049 & 0.2570 & 0.9554 & 0.2743 & 0.4109 & 0.0809 & 0.1031\\
&\!\!-DI& 0.8501 & 0.2966 & 0.4476 & 0.1642 & 0.2039 & 0.9516 & 0.2648 & 0.3952 & 0.0804 & 0.1013\\
&\!\!-DUI& \textbf{0.8743} & \textbf{0.4470} & \textbf{0.5251} & \textbf{0.2946} & \textbf{0.2920} & \textbf{0.9575} & \textbf{0.3027} & \textbf{0.4367} & \textbf{0.0942} & \textbf{0.1158}\\
\hline
\end{tabular}}}
\end{table*}

\subsubsection{Baselines}
We compare our methods with the following state-of-the-art methods:

-\textbf{MF-$\lambda$\textsc{Fix}}. For the fixed regularization methods, we use a global $\lambda$ for all latent dimensions, users and items due to limited computation resources. To simulate how the practitioner would select the best $\lambda$, we searched for the best $\lambda$ among $\{10, 1, 10^{-1}\!, 10^{-2}\!, 10^{-3}\!,10^{-4},$ $ 10^{-5}, 0\}$. For each $\lambda$, we ran a full training of the model and checked its performance on the validation set. The best one is selected for comparison with the model trained by our \textsc{$\lambda$Opt} regularizer.

- \textbf{SGDA} \cite{Rendle2012LearningRegularization}. As mentioned in Section \ref{sec:introduction}, SGDA is an adaptive regularization method based on dimension-wise regularization and SGD, derived for rating prediction task. For top-K recommendation, we have to re-derive it by hand. Luckily, it can be roughly seen as $\lambda$\textsc{Opt} with dimension-wise $\Lambda$ and SGD for MF updates. We find that the implanted SGDA can be very hard to tune due to its limitation in optimizer choices. As SGD doesn't adjust the learning rate during training, the initial learning rate is crucial to the final performance. Meanwhile, it is pretty hard to tune the learning rate for SGD as small learning rate leads to slow convergence and large learning rate might contribute to bad performance. Here we follow the same tuning flow as we tune the models for our approach. We first find an appropriate learning rate for the fixed SGD MF. And then we add the SGDA regularizer to the SGD MF with this learning rate and tune the hyperparameters for the SGDA regularizer.

- \textbf{AMF} \cite{he2018adversarial}. Adversarial Matrix Factorization (AMF) is a state-of-the-art model for recommendation. It employs Adversarial Personalized Ranking (APR) method, which enhances vanilla MF by performing adversarial training. In our experiments, we start adversarial training when the loss of MF-$\lambda$\textsc{Fix} converges and tune the learning rate and $L_2$ regularizer, as described in the paper.

- \textbf{NeuMF} \cite{he2017neural}. Neural Matrix Factorization (NeuMF) is a state-of-the-art neural model for item recommendation, combining MF and Multi-Layer Perceptrons (MLP) to learn user-item interaction. Following \cite{he2017neural}, we first pretrain the model with MF-$\lambda$\textsc{Fix}, and then tune the depth of MLP, learning rate, and $L_2$ regularization.

As for our method, we adopt Adam as the optimizer for MF update. The reason is that, in our experiments, we find the training of recommender models with adaptive regularizers is much more sensitive to the step sizes compared to training with the fixed regularizer. Practitioners of such methods could be eager for optimizers that adapt their learning rates during the training procedure of recommender models. Since SGDA only gives $\lambda$-update solutions to the models optimized by vanilla SGD, our method is more generic in terms of endowing the practitioners more freedom in optimizer choices. For fair comparison, we use the same MF model configurations for MF-$\lambda$\textsc{Fix}, SGDA, AMF and $\lambda$\textsc{Opt} as we want to justify the effect of various regularization strategies.

\subsection{Performance of \textsc{$\lambda$Opt} (RQ.1)}
Table \ref{tab:results} shows the results. Our methods are named as "MF-\textsc{$\lambda$Opt}-[regularization granularity]", with ``D'', ``U'', ``I'' and ``DUI'' standing for dimension-wise, user-wise, item-wise and the three respectively. On both datasets, MF-\textsc{$\lambda$Opt}-DUI outperforms the other methods by a large margin -- about 10\%-20\% in HR and NDCG. The variants of \textsc{$\lambda$Opt} with different regularization granularity also show promising performance, which indicates that \textsc{$\lambda$Opt} lowers down the computation barrier and prerequisite for practitioners without hurting model performance. In fact, it even boosts the recommendation performance if combined with suitable fine-grained regularization.

\subsection{Sparseness and Activeness (RQ.2)}
Section 5.2 demonstrates \textsc{$\lambda$Opt}'s superiority over others. But where does $\lambda$\textsc{Opt} gain its performance improvements? Do they come from handling the sparse users better than the fixed $\lambda$ approach? Or do they come from addressing active users better?  In order to validate $\lambda$\textsc{Opt}'s, to be exact,  MF-\textsc{$\lambda$Opt}-DUI's effectiveness in addressing heterogeneous users/items, we check its performance improvements across users/items with varied frequencies. Due to limited space, we only present figures on Amazon Food Review Dataset. The results on MovieLens 10M are similar.

\begin{figure}[tbp]
\centering
\vspace{-1mm}
\subfigure[Users on Amazon Food Review]{\includegraphics[width=0.47\linewidth]{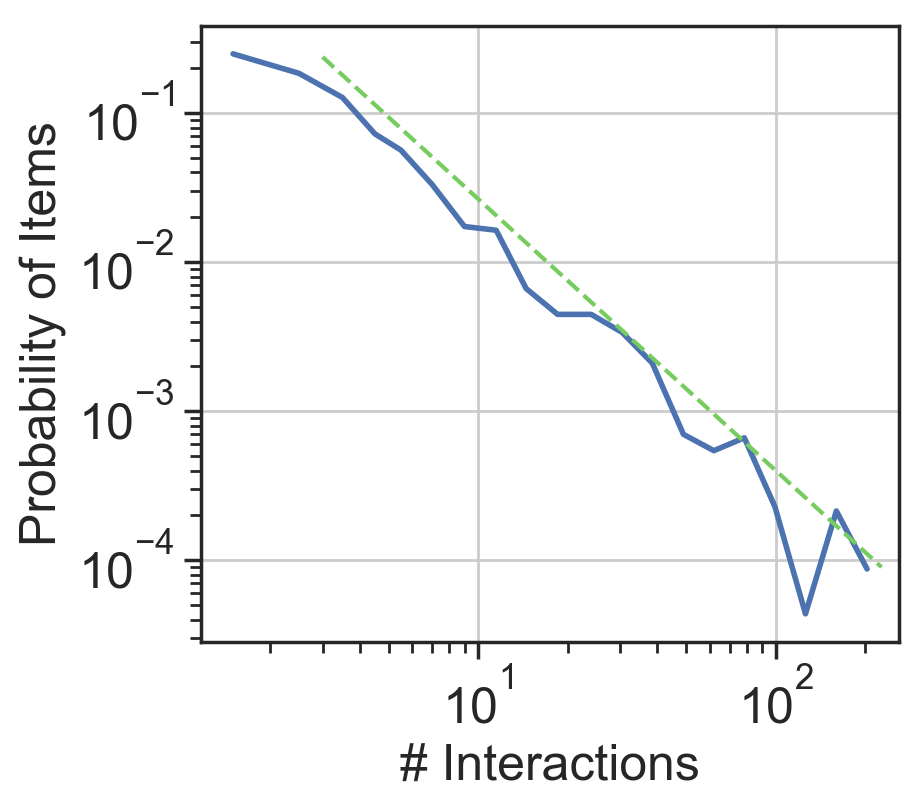}}
\subfigure[Items on Amazon Food Review]{\includegraphics[width=0.47\linewidth]{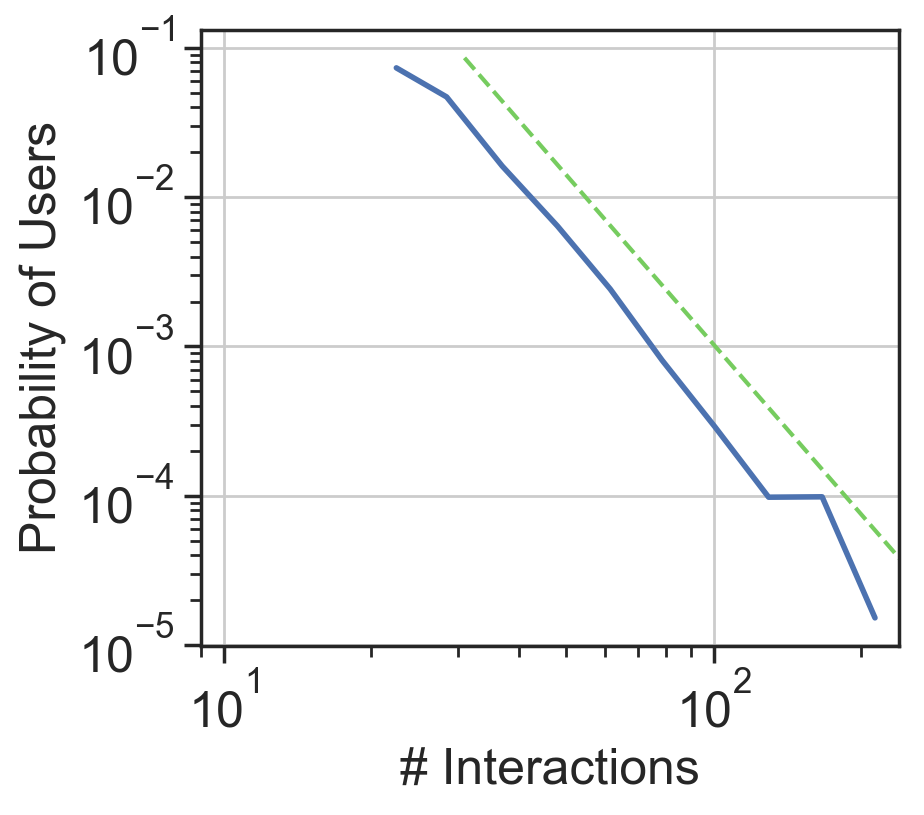}}
\vspace{-2mm}
\caption{Distributions of user and item frequencies.}
\label{fig:longtail}
\end{figure}

Figure \ref{fig:longtail} shows the distributions of user and item frequencies on Amazon Food Review. We can observe that they are all long-tailed. This poses a great challenge to the recommender model, as it needs to be flexible enough to take care of both the head users/items (sparse) and tail users/items (active). Regularization strategies that set a global $\lambda$ or dimension-wise $\lambda$ for all users/items might not work well as they cannot address both types of users. Choosing an appropriate $\lambda$ via grid-search, in essence, seeks compromise between sparse and active users/items. In contrast, regularization in finer levels, e.g. user-/item-aware regularization, obviates the need to compromise between users/items with diverse frequencies. 

\begin{figure*}[tbp]
\centering
\subfigure[HR@100 for users]{\includegraphics[width=0.24\linewidth]{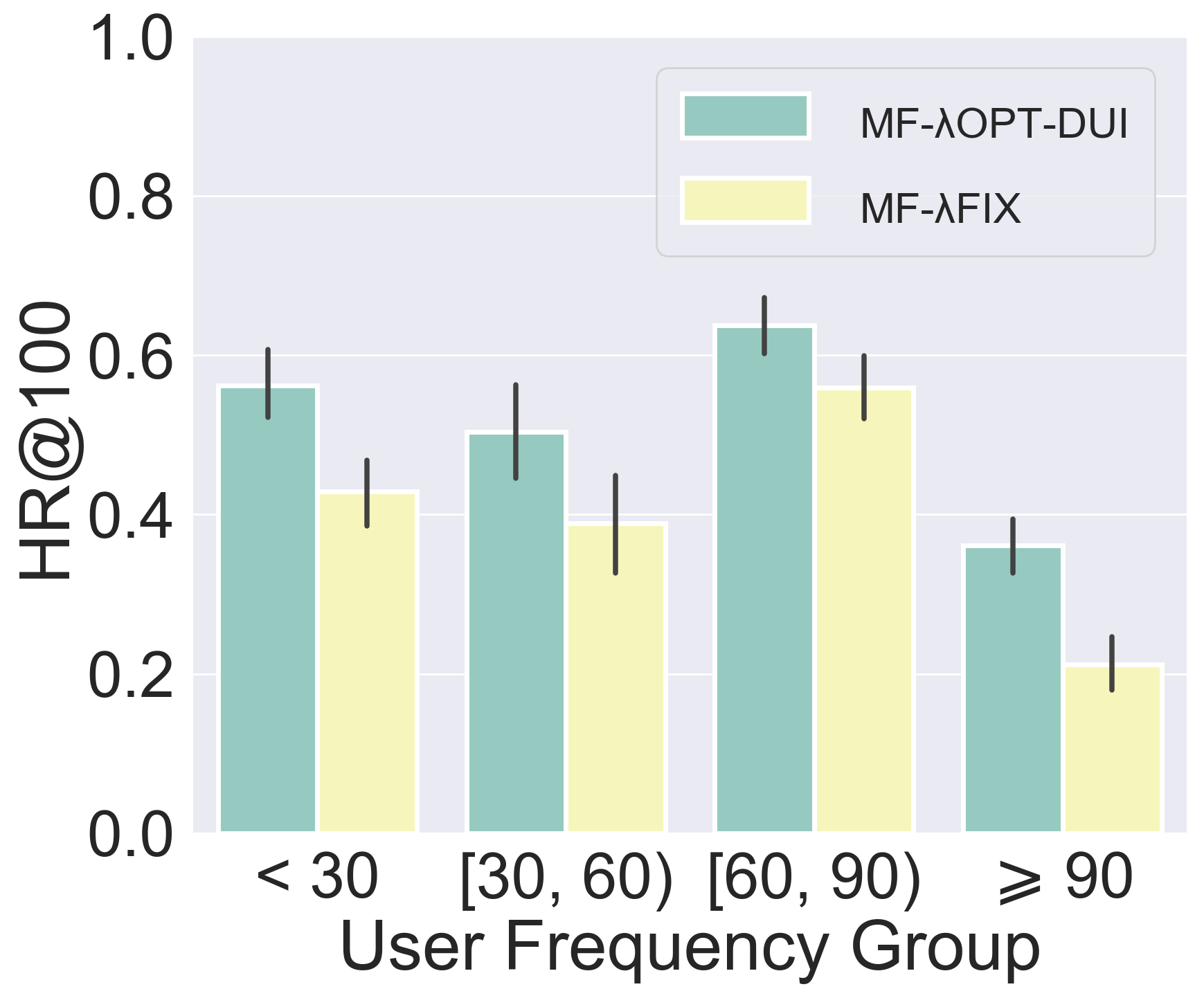}}
\subfigure[NDCG@100 for users]{\includegraphics[width=0.24\linewidth]{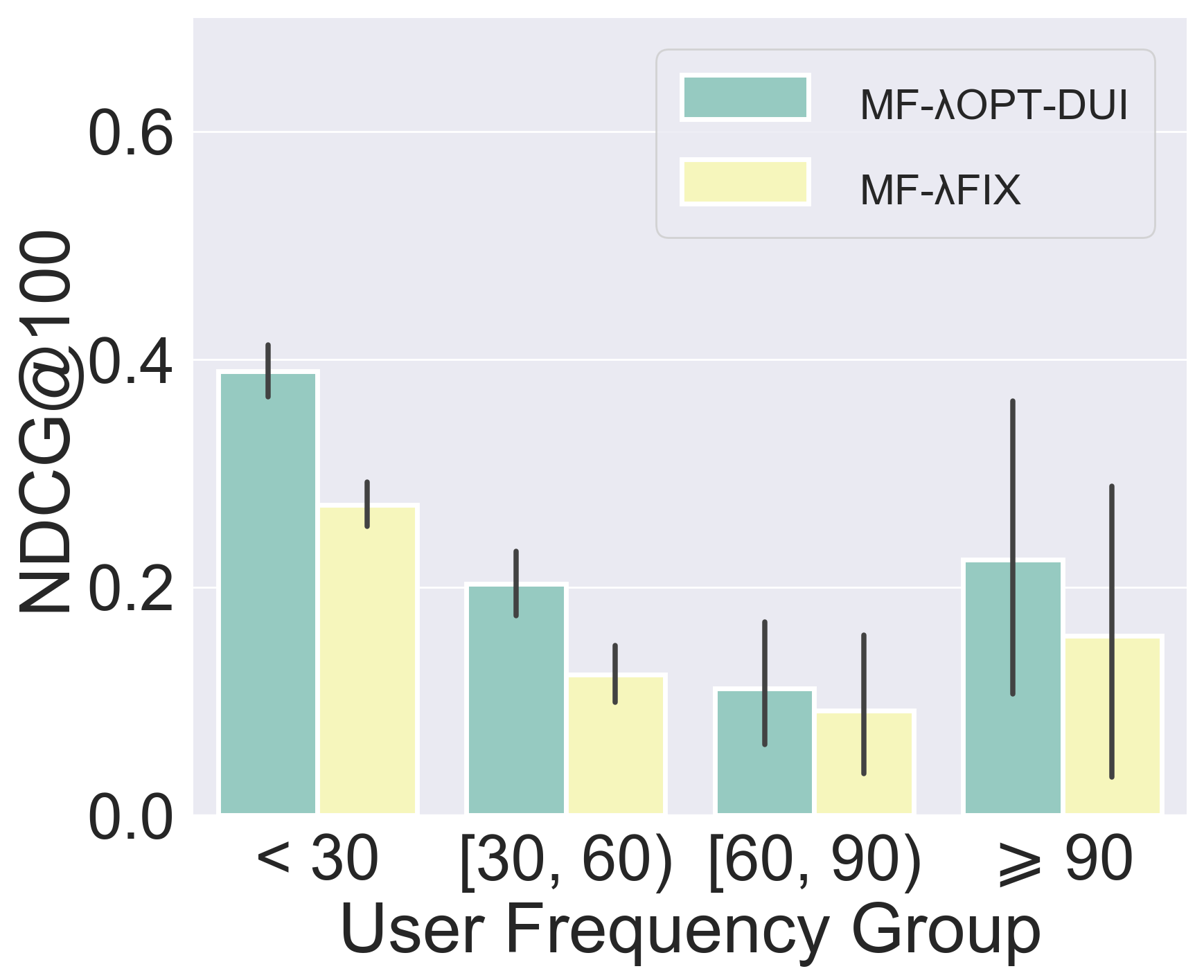}}
\subfigure[HR@100 for items]{\includegraphics[width=0.24\linewidth]{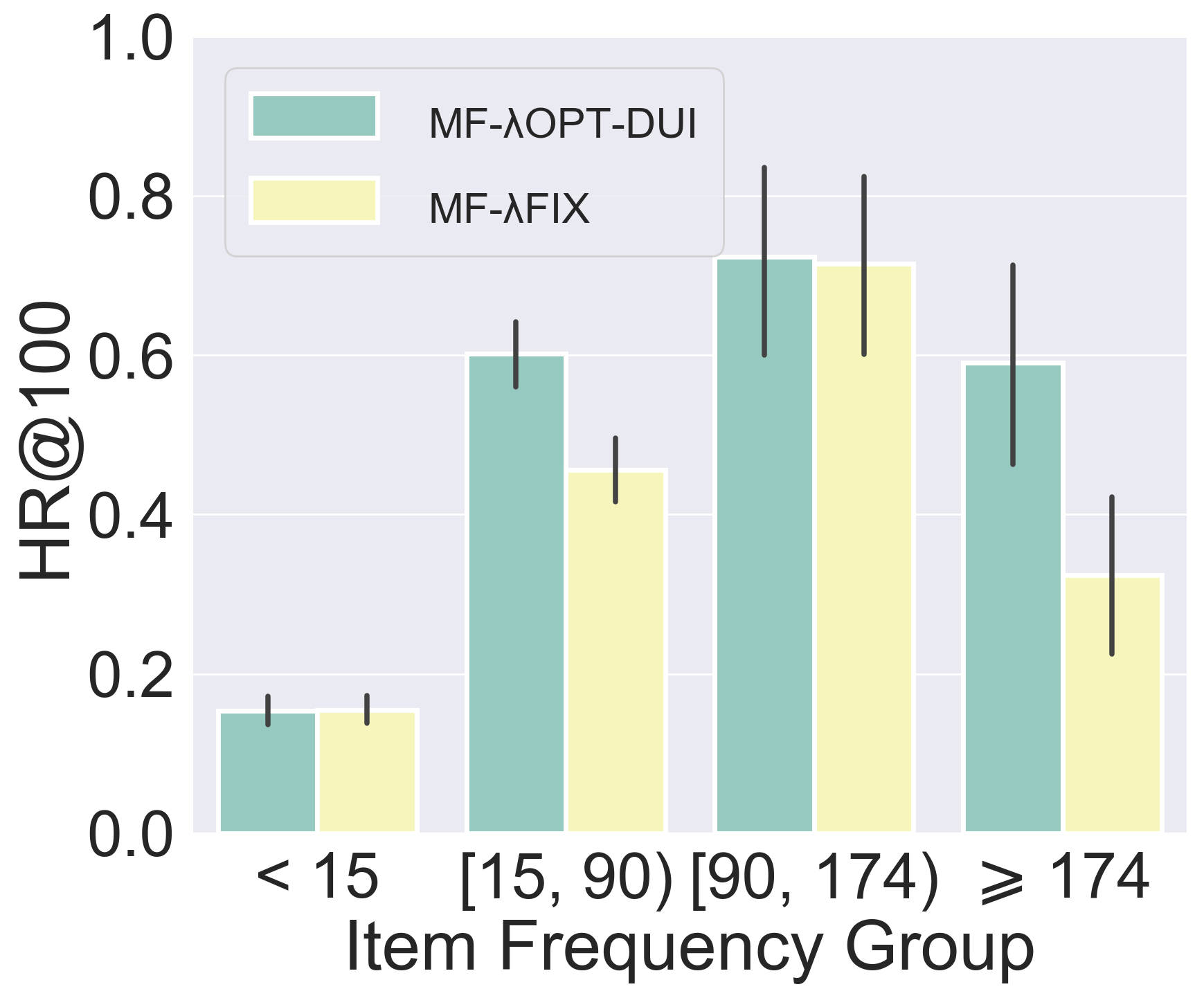}}
\subfigure[NDCG@100 for items]{\includegraphics[width=0.24\linewidth]{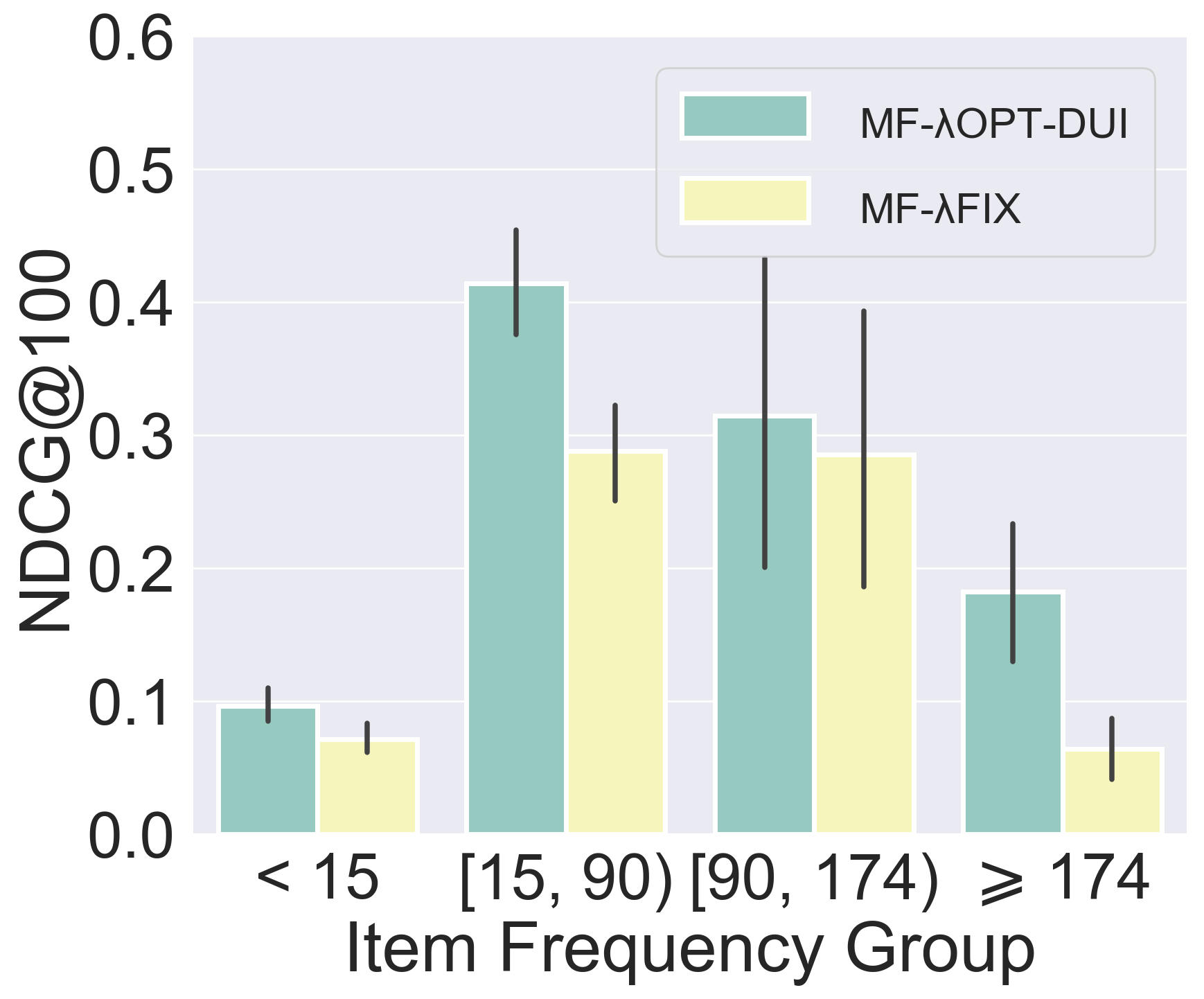}}
\vspace{-4mm}
\caption{Performance improvements of users/items in varied frequency groups on Amazon Food Review.}
\label{fig:perform food}
\vspace{-2mm}
\end{figure*}

MF-$\lambda$\textsc{Opt}-DUI specifies an individual level of regularization strength for each user, item and dimension. We investigate its impact on users/items with varied numbers of interactions. Figure \ref{fig:perform food} shows its performance improvements (HR@100 and NDCG@100) over the fixed approach on Amazon Food Review. The HR and NDCG for users are defined as stated in the experimental settings. For an item, as it is not convenient to compute item-wise measures in BPR setting, we compute its "HR" and "NDCG" as follows: In the test set, we find the users that interact with the item and take the average of their HRs as the "HR" for the item; we find the users that interact with the item and take the average of their NDCGs as the "NDCG" for the item. Such measures are useful in making comparisons across different methods. 

 As we can observe from Figure \ref{fig:perform food}, except <15 item group, MF-$\lambda$\textsc{Opt}-DUI lifts performances by about $10\%$ across users/items in varied frequency groups, conforming to our design that \textsc{$\lambda$Opt} can handle both sparseness and activeness better.

\subsection{Analysis of $\Lambda$-trajectories (RQ.3)}
How does $\lambda$\textsc{Opt} adjust $\lambda$ to gain such surprising improvement across heterogeneous users and items? In other words, does $\lambda$\textsc{Opt} find special $\Lambda$-trajectories for them? We dive into this research question by analyzing the $\Lambda$-trajectories. 

\begin{figure}[tbp]
\centering
\subfigure[For users on Amazon Food Review]{\includegraphics[width=0.485\linewidth]{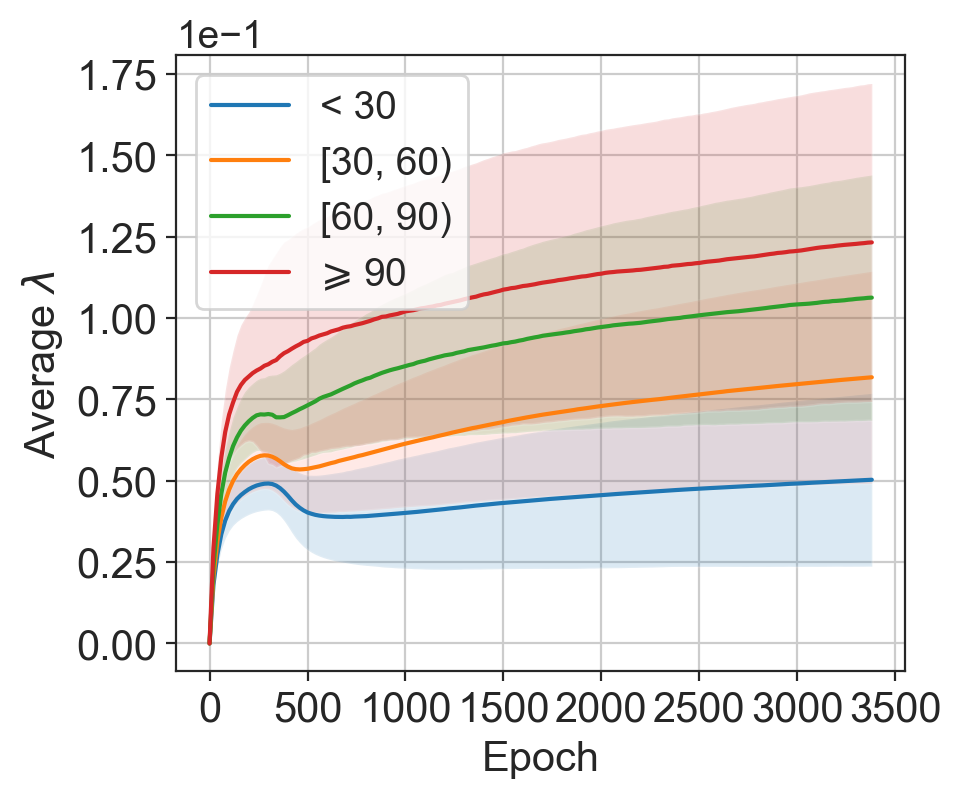}}
\subfigure[For items on Amazon Food Review]{\includegraphics[width=0.46\linewidth]{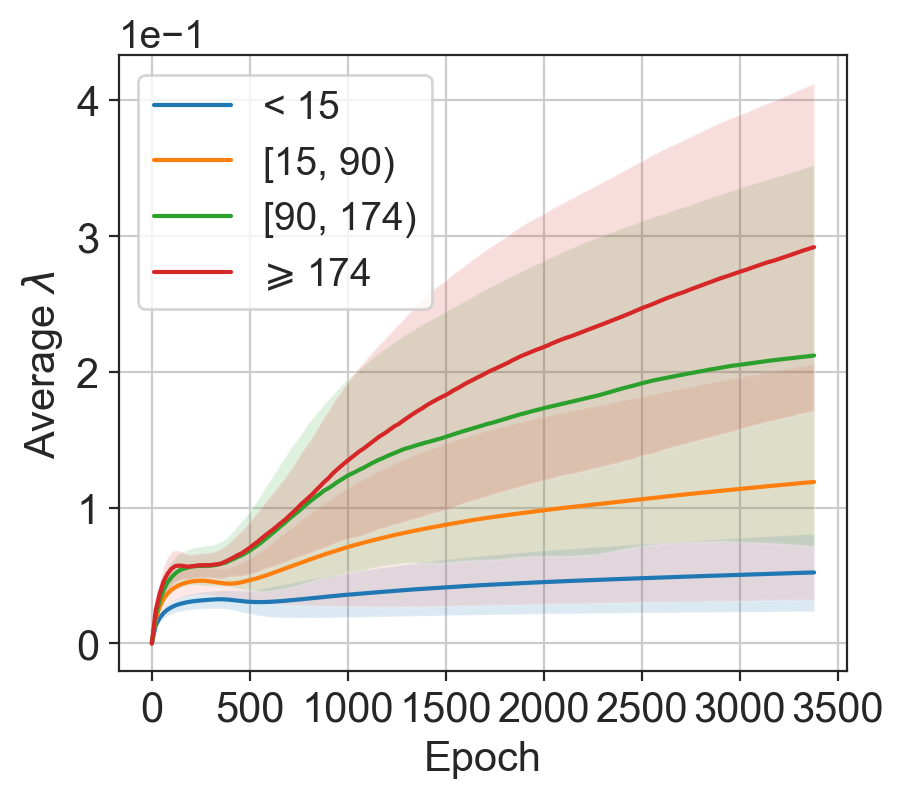}}
\vspace{-2mm}
\caption{$\Lambda$-trajectories generated by MF-$\lambda$\textsc{Opt}-DUI.}
\label{fig:lambda epoch}
\vspace{-2mm}
\end{figure}

\begin{figure}[tbp]
\centering
\subfigure[For users on Amazon Food Review]{\includegraphics[width=0.47\linewidth]{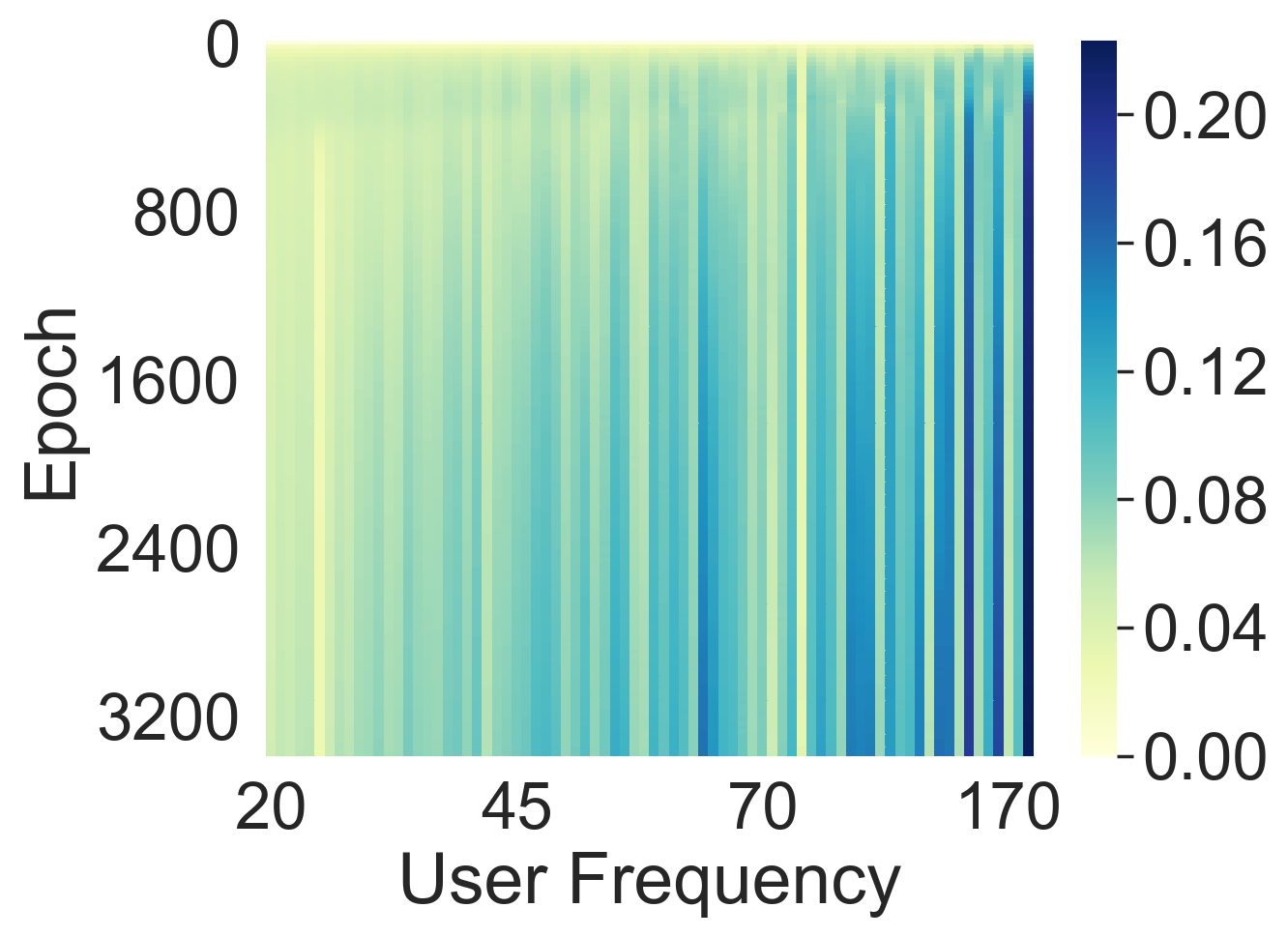}}
\subfigure[For items on Amazon Food Review]{\includegraphics[width=0.45\linewidth]{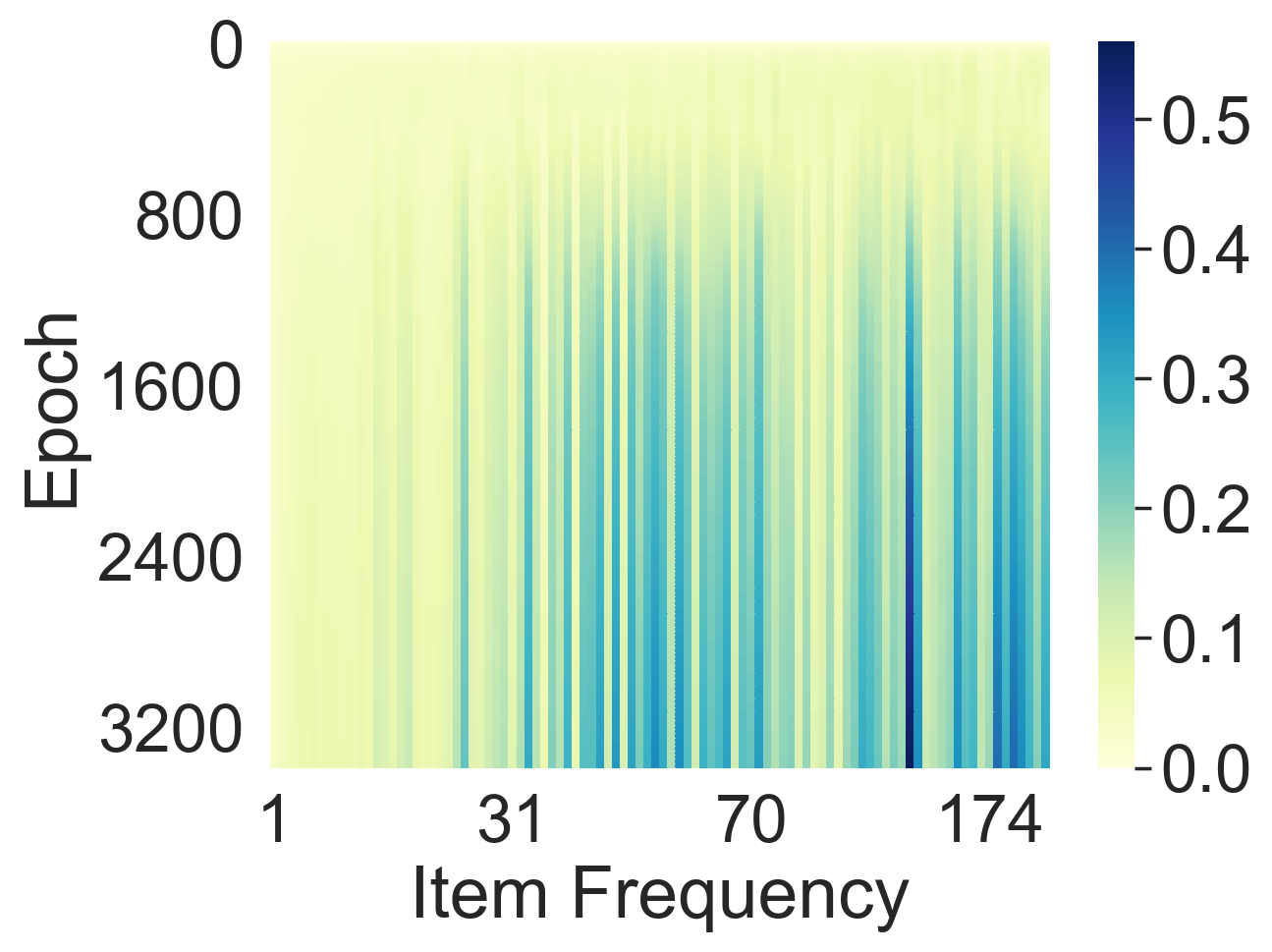}}
\vspace{-3mm}
\caption{Relationship between frequencies and average $\lambda$ by MF-$\lambda$\textsc{Opt}-DUI, evolving as training epoch increases. Color indicates the magnitude of average $\lambda$.}
\label{fig:lambda freq}
\vspace{-2mm}
\end{figure}

Figure \ref{fig:lambda epoch} shows the $\Lambda$-trajectories. Different colors indicate different groups of users and items divided according to numbers of interactions. For every user and item, we aggregate the $\Lambda$ over all the latent dimensions. We then take the average of all users or items within the same group. The variance within a group is indicated by the shades in the figure. Figure \ref{fig:lambda freq} shows the relationship between $\lambda$ and the frequency for users/items, evolving as training epoch increases. Color implies the magnitudes of $\lambda$.

As we could see, most $\lambda$s tend to increase as the training procedure goes on. This conforms to the intuition that regularization is necessary after the model encounters the data multiple times. Difference among groups can also be observed. Interestingly, the active users receive stronger regularization after epochs of training although the initial $\Lambda$ is zero for all users. A possible explanation would be the active users have more data and the model learns from the data so quickly that it might get overfitting to them, making strong regularization necessary. Under these circumstances, a global strong regularization level would satisfy the active users but at the risk of failing the sparse users. In contrast, $\lambda$\textsc{Opt} finds a special $\Lambda$-trajectory for every user, being capable to please both sparse users and active users. As shown in Figure \ref{fig:lambda epoch}, sparse users receive a relatively weak regularization so that recommendation to them could rely more on the data. The findings on the item $\Lambda$-trajectories are similar. We could conclude here that $\lambda$\textsc{Opt} owns most of its improvements to finding "personalized" $\Lambda$-trajectories for a diverse set of users and items. We believe this property is valuable to most large-scale web applications where long-tailed phenomenons are common and data sparsity remains a severe challenge. 

So far, we have justified that $\lambda$\textsc{Opt} not only lowers down the computation cost needed to search for a good regularization level but also has the potential to boost the recommendation performance by fine-grained regularization. We also reveal its secret in improving the recommendation performance by analyzing the $\Lambda$-trajectories.

\section{Related Work}
\subsubsection*{\textbf{Adaptive Regularization for Rating Prediction}.} The close work is SGDA \cite{Rendle2012LearningRegularization}, where adaptive regularization for rating prediction is achieved by alternating optimization for model parameters $\Theta$ and regularization coefficients $\lambda$. Similar validation set based alternating optimization method has also been proposed in \cite{Luketina2016ScalableHyperparameters}. Both work focused on the reduced computation complexity while ignoring the potential performance boost. \textbf{As far as we know, we are the first to reveal the important insight that, adaptive regularization in finer levels can bring additional performance benefits for recommender systems.} 
\cite{Rendle2012LearningRegularization} only considers dimension-wise $\lambda$, which might be the reason why the algorithm does not outperform the best fixed $\lambda$ algorithm in the reported experimental results. 
Instead, our work shows the effectiveness of incorporating fine-grained regularization. Besides, our method is more generic in terms of endowing the practitioners more freedom in optimizer choices while SGDA\cite{Rendle2012LearningRegularization} applies only to SGD optimizers. 

\subsubsection*{\textbf{Hyperparameters Optimization}.} 
Finding good regularization coefficients can be part of the overall hyperparameters optimization (HO). Typically, grid-search-like methods are used where people monitor performance on the validation set and choose the best set of hyperparameters from a bunch of candidates. These methods are simple and generic, capable of being applied to any task and any model, ranging from SVM \cite{hsu2003practical} to decision trees. Random search could be very time-consuming. Previous work \cite{snoek2012practical, shahriari2016taking, snoek2015scalable, li2016efficient, feurer2014using} have dedicated to lower down the nontrivial search cost, along with developing some enhanced toolboxes \cite{martinez2014bayesopt,bergstra2013hyperopt,claesen2014easy}.
However, most of them require multiple full training runs instead of learning to regularize on the fly. Recently, \cite{franceschi2018bilevel} explored bilevel programming to unify gradient based HO and meta-learning.
\textbf{The above hyper-parameters optimization methods do not specialize on recommender systems}. Applying them to tuning the regularization coefficients for recommendation might not work well due to some characteristics, i.e., data sparsity issue, in recommender systems. In contrast, our algorithms are tailored to recommendation, where users/items are highly heterogeneous.

\subsubsection*{\textbf{Regularization of Embeddings}.} 
Embedding technique is widely used to project categorical values into a latent vector space \cite{guo2016entity}. In natural language processing, training large embeddings usually requires suitable regularization \cite{peng2015comparative}. Training recommender models also involves regularizing large embedding matrices, such as the user/item embedding matrix. Although the tasks are different, the basic regularization strategies and their analysis might be similar. A cross-sectional study across them would be interesting and meaningful in terms of deriving a generic regularization method for embeddings. Since parameters initialization can be regarded as a special regularization, embedding initialization methods like \cite{pan2019warm} are also worth exploring.

\section{Conclusion and Future Work}
Tuning regularization hyperparameters for recommender models has been a tedious even miserable job for practitioners. We propose a generic method, \textsc{$\lambda$Opt}, to address this problem. \textsc{$\lambda$Opt} adjusts the regularization hyperparameters on the fly based on validation data. Our experiments based on two benchmarks demonstrate that \textsc{$\lambda$Opt} can be a simple yet effective training tool in terms of lower computation cost and better performance with fine-grained regularization. 

In future, we are interested in extending our method to more complex models, such as FM and NeuMF. Since \textsc{$\lambda$Opt} relies on validation data to update the regularization coefficients, it requires sufficient data to ensure the generalization \cite{Rendle2012LearningRegularization, snoek2015scalable} when the number of hyperparameters is large. It would be worthwhile to investigate the influence of validation data size. Moreover, we would like to dive into the theoretical foundation for such adaptive hyperparameter optimization methods, which is significant for further applications.  

%
\begin{acks}
This work is partly supported by the National Natural Science Foundation of China (Grant No. 61673237, 61621091 and 61861136003), Beijing National Research Center for Information Science and Technology under 20031887521, and the Thousand Youth Talents Program 2018. Thank Kun Yan for his help with figure drawing. Thank reviewers for their constructive comments.
\end{acks}

%
\bibliographystyle{ACM-Reference-Format}
\bibliography{reference-regularize}


\begin{thebibliography}{33}


\ifx \showCODEN    \undefined \def \showCODEN     #1{\unskip}     \fi
\ifx \showDOI      \undefined \def \showDOI       #1{#1}\fi
\ifx \showISBNx    \undefined \def \showISBNx     #1{\unskip}     \fi
\ifx \showISBNxiii \undefined \def \showISBNxiii  #1{\unskip}     \fi
\ifx \showISSN     \undefined \def \showISSN      #1{\unskip}     \fi
\ifx \showLCCN     \undefined \def \showLCCN      #1{\unskip}     \fi
\ifx \shownote     \undefined \def \shownote      #1{#1}          \fi
\ifx \showarticletitle \undefined \def \showarticletitle #1{#1}   \fi
\ifx \showURL      \undefined \def \showURL       {\relax}        \fi
\providecommand\bibfield[2]{#2}
\providecommand\bibinfo[2]{#2}
\providecommand\natexlab[1]{#1}
\providecommand\showeprint[2][]{arXiv:#2}

\bibitem[\protect\citeauthoryear{Abadi, Barham, Chen, Chen, Davis, Dean, Devin,
  Ghemawat, Irving, Isard, et~al\mbox{.}}{Abadi et~al\mbox{.}}{2016}]%
        {abadi2016tensorflow}
\bibfield{author}{\bibinfo{person}{Mart{\' \i}n Abadi}, \bibinfo{person}{Paul
  Barham}, \bibinfo{person}{Jianmin Chen}, \bibinfo{person}{Zhifeng Chen},
  \bibinfo{person}{Andy Davis}, \bibinfo{person}{Jeffrey Dean},
  \bibinfo{person}{Matthieu Devin}, \bibinfo{person}{Sanjay Ghemawat},
  \bibinfo{person}{Geoffrey Irving}, \bibinfo{person}{Michael Isard},
  {et~al\mbox{.}}} \bibinfo{year}{2016}\natexlab{}.
\newblock \showarticletitle{Tensorflow: a system for large-scale machine
  learning.}
\newblock


\bibitem[\protect\citeauthoryear{Agarwal and Chen}{Agarwal and Chen}{2011}]%
        {tutorial2011}
\bibfield{author}{\bibinfo{person}{Deepak Agarwal} {and}
  \bibinfo{person}{Bee-Chung Chen}.} \bibinfo{year}{2011}\natexlab{}.
\newblock \bibinfo{booktitle}{\emph{Machine Learning for Large Scale
  Recommender Systems}}.
\newblock
\newblock
\shownote{\url{http://pages.cs.wisc.edu/~beechung/icml11-tutorial/}.}


\bibitem[\protect\citeauthoryear{Bayer, He, Kanagal, and Rendle}{Bayer
  et~al\mbox{.}}{2017}]%
        {bayer2017generic}
\bibfield{author}{\bibinfo{person}{Immanuel Bayer}, \bibinfo{person}{Xiangnan
  He}, \bibinfo{person}{Bhargav Kanagal}, {and} \bibinfo{person}{Steffen
  Rendle}.} \bibinfo{year}{2017}\natexlab{}.
\newblock \showarticletitle{A generic coordinate descent framework for learning
  from implicit feedback}. In \bibinfo{booktitle}{\emph{WWW}}.
\newblock


\bibitem[\protect\citeauthoryear{Bergstra, Yamins, and Cox}{Bergstra
  et~al\mbox{.}}{2013}]%
        {bergstra2013hyperopt}
\bibfield{author}{\bibinfo{person}{James Bergstra}, \bibinfo{person}{Dan
  Yamins}, {and} \bibinfo{person}{David~D Cox}.}
  \bibinfo{year}{2013}\natexlab{}.
\newblock \showarticletitle{Hyperopt: A python library for optimizing the
  hyperparameters of machine learning algorithms}. In
  \bibinfo{booktitle}{\emph{Python in Science Conference}}.
  \bibinfo{pages}{13--20}.
\newblock


\bibitem[\protect\citeauthoryear{Chapelle, Vapnik, Bousquet, and
  Mukherjee}{Chapelle et~al\mbox{.}}{2002}]%
        {chapelle2002choosing}
\bibfield{author}{\bibinfo{person}{Olivier Chapelle}, \bibinfo{person}{Vladimir
  Vapnik}, \bibinfo{person}{Olivier Bousquet}, {and} \bibinfo{person}{Sayan
  Mukherjee}.} \bibinfo{year}{2002}\natexlab{}.
\newblock \showarticletitle{Choosing multiple parameters for support vector
  machines}.
\newblock \bibinfo{journal}{\emph{Machine learning}} \bibinfo{volume}{46},
  \bibinfo{number}{1-3} (\bibinfo{year}{2002}), \bibinfo{pages}{131--159}.
\newblock


\bibitem[\protect\citeauthoryear{Chen, Zhang, He, Nie, Liu, and Chua}{Chen
  et~al\mbox{.}}{2017}]%
        {chen2017attentive}
\bibfield{author}{\bibinfo{person}{Jingyuan Chen}, \bibinfo{person}{Hanwang
  Zhang}, \bibinfo{person}{Xiangnan He}, \bibinfo{person}{Liqiang Nie},
  \bibinfo{person}{Wei Liu}, {and} \bibinfo{person}{Tat-Seng Chua}.}
  \bibinfo{year}{2017}\natexlab{}.
\newblock \showarticletitle{Attentive collaborative filtering: Multimedia
  recommendation with item-and component-level attention}. In
  \bibinfo{booktitle}{\emph{SIGIR}}.
\newblock


\bibitem[\protect\citeauthoryear{Claesen, Simm, Popovic, Moreau, and
  De~Moor}{Claesen et~al\mbox{.}}{2014}]%
        {claesen2014easy}
\bibfield{author}{\bibinfo{person}{Marc Claesen}, \bibinfo{person}{Jaak Simm},
  \bibinfo{person}{Dusan Popovic}, \bibinfo{person}{Yves Moreau}, {and}
  \bibinfo{person}{Bart De~Moor}.} \bibinfo{year}{2014}\natexlab{}.
\newblock \showarticletitle{Easy hyperparameter search using Optunity}.
\newblock \bibinfo{journal}{\emph{arXiv:1412.1114}} (\bibinfo{year}{2014}).
\newblock


\bibitem[\protect\citeauthoryear{Cremonesi, Koren, and Turrin}{Cremonesi
  et~al\mbox{.}}{2010}]%
        {Cremonesi:2010}
\bibfield{author}{\bibinfo{person}{Paolo Cremonesi}, \bibinfo{person}{Yehuda
  Koren}, {and} \bibinfo{person}{Roberto Turrin}.}
  \bibinfo{year}{2010}\natexlab{}.
\newblock \showarticletitle{Performance of Recommender Algorithms on Top-n
  Recommendation Tasks}. In \bibinfo{booktitle}{\emph{RecSys}}.
\newblock


\bibitem[\protect\citeauthoryear{Duchi, Hazan, and Singer}{Duchi
  et~al\mbox{.}}{2011}]%
        {duchi2011adaptive}
\bibfield{author}{\bibinfo{person}{John Duchi}, \bibinfo{person}{Elad Hazan},
  {and} \bibinfo{person}{Yoram Singer}.} \bibinfo{year}{2011}\natexlab{}.
\newblock \showarticletitle{Adaptive subgradient methods for online learning
  and stochastic optimization}.
\newblock \bibinfo{journal}{\emph{JMLR}} \bibinfo{volume}{12},
  \bibinfo{number}{Jul} (\bibinfo{year}{2011}), \bibinfo{pages}{2121--2159}.
\newblock


\bibitem[\protect\citeauthoryear{Dyer, Owen, et~al\mbox{.}}{Dyer
  et~al\mbox{.}}{2011}]%
        {dyer2011visualizing}
\bibfield{author}{\bibinfo{person}{Justin~S Dyer}, \bibinfo{person}{Art~B
  Owen}, {et~al\mbox{.}}} \bibinfo{year}{2011}\natexlab{}.
\newblock \showarticletitle{Visualizing bivariate long-tailed data}.
\newblock \bibinfo{journal}{\emph{Electronic Journal of Statistics}}
  \bibinfo{volume}{5} (\bibinfo{year}{2011}), \bibinfo{pages}{642--668}.
\newblock


\bibitem[\protect\citeauthoryear{Feurer, Springenberg, and Hutter}{Feurer
  et~al\mbox{.}}{2014}]%
        {feurer2014using}
\bibfield{author}{\bibinfo{person}{Matthias Feurer},
  \bibinfo{person}{Jost~Tobias Springenberg}, {and} \bibinfo{person}{Frank
  Hutter}.} \bibinfo{year}{2014}\natexlab{}.
\newblock \showarticletitle{Using meta-learning to initialize bayesian
  optimization of hyperparameters}. In \bibinfo{booktitle}{\emph{International
  Conference on Meta-learning and Algorithm Selection}}.
\newblock


\bibitem[\protect\citeauthoryear{Franceschi, Frasconi, Salzo, Grazzi, and
  Pontil}{Franceschi et~al\mbox{.}}{2018}]%
        {franceschi2018bilevel}
\bibfield{author}{\bibinfo{person}{Luca Franceschi}, \bibinfo{person}{Paolo
  Frasconi}, \bibinfo{person}{Saverio Salzo}, \bibinfo{person}{Riccardo
  Grazzi}, {and} \bibinfo{person}{Massimiliano Pontil}.}
  \bibinfo{year}{2018}\natexlab{}.
\newblock \showarticletitle{Bilevel Programming for Hyperparameter Optimization
  and Meta-Learning}. In \bibinfo{booktitle}{\emph{International Conference on
  Machine Learning}}. \bibinfo{pages}{1563--1572}.
\newblock


\bibitem[\protect\citeauthoryear{Guo and Berkhahn}{Guo and Berkhahn}{2016}]%
        {guo2016entity}
\bibfield{author}{\bibinfo{person}{Cheng Guo} {and} \bibinfo{person}{Felix
  Berkhahn}.} \bibinfo{year}{2016}\natexlab{}.
\newblock \showarticletitle{Entity embeddings of categorical variables}.
\newblock \bibinfo{journal}{\emph{arXiv:1604.06737}} (\bibinfo{year}{2016}).
\newblock


\bibitem[\protect\citeauthoryear{He and Chua}{He and Chua}{2017}]%
        {NFM}
\bibfield{author}{\bibinfo{person}{Xiangnan He} {and} \bibinfo{person}{Tat-Seng
  Chua}.} \bibinfo{year}{2017}\natexlab{}.
\newblock \showarticletitle{Neural factorization machines for sparse predictive
  analytics}. In \bibinfo{booktitle}{\emph{SIGIR}}.
\newblock


\bibitem[\protect\citeauthoryear{He, He, Du, and Chua}{He
  et~al\mbox{.}}{2018}]%
        {he2018adversarial}
\bibfield{author}{\bibinfo{person}{Xiangnan He}, \bibinfo{person}{Zhankui He},
  \bibinfo{person}{Xiaoyu Du}, {and} \bibinfo{person}{Tat-Seng Chua}.}
  \bibinfo{year}{2018}\natexlab{}.
\newblock \showarticletitle{Adversarial personalized ranking for
  recommendation}. In \bibinfo{booktitle}{\emph{SIGIR}}.
\newblock


\bibitem[\protect\citeauthoryear{He, Liao, Zhang, Nie, Hu, and Chua}{He
  et~al\mbox{.}}{2017}]%
        {he2017neural}
\bibfield{author}{\bibinfo{person}{Xiangnan He}, \bibinfo{person}{Lizi Liao},
  \bibinfo{person}{Hanwang Zhang}, \bibinfo{person}{Liqiang Nie},
  \bibinfo{person}{Xia Hu}, {and} \bibinfo{person}{Tat-Seng Chua}.}
  \bibinfo{year}{2017}\natexlab{}.
\newblock \showarticletitle{Neural collaborative filtering}. In
  \bibinfo{booktitle}{\emph{WWW}}.
\newblock


\bibitem[\protect\citeauthoryear{Hsu, Chang, Lin, et~al\mbox{.}}{Hsu
  et~al\mbox{.}}{2003}]%
        {hsu2003practical}
\bibfield{author}{\bibinfo{person}{Chih-Wei Hsu}, \bibinfo{person}{Chih-Chung
  Chang}, \bibinfo{person}{Chih-Jen Lin}, {et~al\mbox{.}}}
  \bibinfo{year}{2003}\natexlab{}.
\newblock \showarticletitle{A practical guide to support vector
  classification}.
\newblock  (\bibinfo{year}{2003}).
\newblock


\bibitem[\protect\citeauthoryear{Kingma and Ba}{Kingma and Ba}{2015}]%
        {kingma2014adam}
\bibfield{author}{\bibinfo{person}{Diederik~P Kingma} {and}
  \bibinfo{person}{Jimmy Ba}.} \bibinfo{year}{2015}\natexlab{}.
\newblock \showarticletitle{Adam: A method for stochastic optimization}.
\newblock \bibinfo{journal}{\emph{ICLR}}.
\newblock


\bibitem[\protect\citeauthoryear{Koren, Bell, and Volinsky}{Koren
  et~al\mbox{.}}{2009}]%
        {koren2009matrix}
\bibfield{author}{\bibinfo{person}{Yehuda Koren}, \bibinfo{person}{Robert
  Bell}, {and} \bibinfo{person}{Chris Volinsky}.}
  \bibinfo{year}{2009}\natexlab{}.
\newblock \showarticletitle{Matrix factorization techniques for recommender
  systems}.
\newblock \bibinfo{journal}{\emph{Computer}} \bibinfo{number}{8}
  (\bibinfo{year}{2009}), \bibinfo{pages}{30--37}.
\newblock


\bibitem[\protect\citeauthoryear{Larsen, Svarer, Andersen, and Hansen}{Larsen
  et~al\mbox{.}}{1998}]%
        {larsen1998adaptive}
\bibfield{author}{\bibinfo{person}{Jan Larsen}, \bibinfo{person}{Claus Svarer},
  \bibinfo{person}{Lars~Nonboe Andersen}, {and} \bibinfo{person}{Lars~Kai
  Hansen}.} \bibinfo{year}{1998}\natexlab{}.
\newblock \showarticletitle{Adaptive regularization in neural network
  modeling}.
\newblock In \bibinfo{booktitle}{\emph{Neural Networks: Tricks of the Trade}}.
  \bibinfo{pages}{113--132}.
\newblock


\bibitem[\protect\citeauthoryear{Li, Jamieson, DeSalvo, Rostamizadeh, and
  Talwalkar}{Li et~al\mbox{.}}{2016}]%
        {li2016efficient}
\bibfield{author}{\bibinfo{person}{Lisha Li}, \bibinfo{person}{Kevin Jamieson},
  \bibinfo{person}{Giulia DeSalvo}, \bibinfo{person}{Afshin Rostamizadeh},
  {and} \bibinfo{person}{Ameet Talwalkar}.} \bibinfo{year}{2016}\natexlab{}.
\newblock \showarticletitle{Efficient hyperparameter optimization and
  infinitely many armed bandits}.
\newblock \bibinfo{journal}{\emph{CoRR, abs/1603.06560}}
  (\bibinfo{year}{2016}).
\newblock


\bibitem[\protect\citeauthoryear{Luketina, Berglund, Greff, and Raiko}{Luketina
  et~al\mbox{.}}{2016}]%
        {Luketina2016ScalableHyperparameters}
\bibfield{author}{\bibinfo{person}{Jelena Luketina}, \bibinfo{person}{Mathias
  Berglund}, \bibinfo{person}{Klaus Greff}, {and} \bibinfo{person}{Tapani
  Raiko}.} \bibinfo{year}{2016}\natexlab{}.
\newblock \showarticletitle{Scalable gradient-based tuning of continuous
  regularization hyperparameters}. In \bibinfo{booktitle}{\emph{ICML}}.
\newblock


\bibitem[\protect\citeauthoryear{Martinez-Cantin}{Martinez-Cantin}{2014}]%
        {martinez2014bayesopt}
\bibfield{author}{\bibinfo{person}{Ruben Martinez-Cantin}.}
  \bibinfo{year}{2014}\natexlab{}.
\newblock \showarticletitle{Bayesopt: A bayesian optimization library for
  nonlinear optimization, experimental design and bandits}.
\newblock \bibinfo{journal}{\emph{JMLR}} \bibinfo{volume}{15},
  \bibinfo{number}{1} (\bibinfo{year}{2014}), \bibinfo{pages}{3735--3739}.
\newblock


\bibitem[\protect\citeauthoryear{Pan, Li, Ao, Tang, and He}{Pan
  et~al\mbox{.}}{2019}]%
        {pan2019warm}
\bibfield{author}{\bibinfo{person}{Feiyang Pan}, \bibinfo{person}{Shuokai Li},
  \bibinfo{person}{Xiang Ao}, \bibinfo{person}{Pingzhong Tang}, {and}
  \bibinfo{person}{Qing He}.} \bibinfo{year}{2019}\natexlab{}.
\newblock \showarticletitle{Warm Up Cold-start Advertisements: Improving CTR
  Predictions via Learning to Learn ID Embeddings}. In
  \bibinfo{booktitle}{\emph{Proceedings of the 42nd annual international ACM
  SIGIR conference on Research and development in information retrieval}}. ACM.
\newblock


\bibitem[\protect\citeauthoryear{Peng, Mou, Li, Chen, Lu, and Jin}{Peng
  et~al\mbox{.}}{2015}]%
        {peng2015comparative}
\bibfield{author}{\bibinfo{person}{Hao Peng}, \bibinfo{person}{Lili Mou},
  \bibinfo{person}{Ge Li}, \bibinfo{person}{Yunchuan Chen},
  \bibinfo{person}{Yangyang Lu}, {and} \bibinfo{person}{Zhi Jin}.}
  \bibinfo{year}{2015}\natexlab{}.
\newblock \showarticletitle{A Comparative Study on Regularization Strategies
  for Embedding-based Neural Networks}. In \bibinfo{booktitle}{\emph{EMNLP}}.
\newblock


\bibitem[\protect\citeauthoryear{Rendle}{Rendle}{2012}]%
        {Rendle2012LearningRegularization}
\bibfield{author}{\bibinfo{person}{Steffen Rendle}.}
  \bibinfo{year}{2012}\natexlab{}.
\newblock \showarticletitle{Learning recommender systems with adaptive
  regularization}. In \bibinfo{booktitle}{\emph{WSDM}}.
\newblock


\bibitem[\protect\citeauthoryear{Rendle, Freudenthaler, Gantner, and
  Schmidt-Thieme}{Rendle et~al\mbox{.}}{2009}]%
        {RendleBPR:Feedback}
\bibfield{author}{\bibinfo{person}{Steffen Rendle}, \bibinfo{person}{Christoph
  Freudenthaler}, \bibinfo{person}{Zeno Gantner}, {and} \bibinfo{person}{Lars
  Schmidt-Thieme}.} \bibinfo{year}{2009}\natexlab{}.
\newblock \showarticletitle{BPR: Bayesian personalized ranking from implicit
  feedback}. In \bibinfo{booktitle}{\emph{UAI}}.
\newblock


\bibitem[\protect\citeauthoryear{Shahriari, Swersky, Wang, Adams, and
  De~Freitas}{Shahriari et~al\mbox{.}}{2016}]%
        {shahriari2016taking}
\bibfield{author}{\bibinfo{person}{Bobak Shahriari}, \bibinfo{person}{Kevin
  Swersky}, \bibinfo{person}{Ziyu Wang}, \bibinfo{person}{Ryan~P Adams}, {and}
  \bibinfo{person}{Nando De~Freitas}.} \bibinfo{year}{2016}\natexlab{}.
\newblock \showarticletitle{Taking the human out of the loop: A review of
  bayesian optimization}.
\newblock \bibinfo{journal}{\emph{IEEE}} \bibinfo{volume}{104},
  \bibinfo{number}{1} (\bibinfo{year}{2016}), \bibinfo{pages}{148--175}.
\newblock


\bibitem[\protect\citeauthoryear{Sinha, Malo, and Deb}{Sinha
  et~al\mbox{.}}{2018}]%
        {sinha2018review}
\bibfield{author}{\bibinfo{person}{Ankur Sinha}, \bibinfo{person}{Pekka Malo},
  {and} \bibinfo{person}{Kalyanmoy Deb}.} \bibinfo{year}{2018}\natexlab{}.
\newblock \showarticletitle{A review on bilevel optimization: from classical to
  evolutionary approaches and applications}.
\newblock \bibinfo{journal}{\emph{IEEE Transactions on Evolutionary
  Computation}} \bibinfo{volume}{22}, \bibinfo{number}{2}
  (\bibinfo{year}{2018}), \bibinfo{pages}{276--295}.
\newblock


\bibitem[\protect\citeauthoryear{Snoek, Larochelle, and Adams}{Snoek
  et~al\mbox{.}}{2012}]%
        {snoek2012practical}
\bibfield{author}{\bibinfo{person}{Jasper Snoek}, \bibinfo{person}{Hugo
  Larochelle}, {and} \bibinfo{person}{Ryan~P Adams}.}
  \bibinfo{year}{2012}\natexlab{}.
\newblock \showarticletitle{Practical bayesian optimization of machine learning
  algorithms}. In \bibinfo{booktitle}{\emph{NIPS}}.
\newblock


\bibitem[\protect\citeauthoryear{Snoek, Rippel, Swersky, Kiros, Satish,
  Sundaram, Patwary, Prabhat, and Adams}{Snoek et~al\mbox{.}}{2015}]%
        {snoek2015scalable}
\bibfield{author}{\bibinfo{person}{Jasper Snoek}, \bibinfo{person}{Oren
  Rippel}, \bibinfo{person}{Kevin Swersky}, \bibinfo{person}{Ryan Kiros},
  \bibinfo{person}{Nadathur Satish}, \bibinfo{person}{Narayanan Sundaram},
  \bibinfo{person}{Mostofa Patwary}, \bibinfo{person}{Mr Prabhat}, {and}
  \bibinfo{person}{Ryan Adams}.} \bibinfo{year}{2015}\natexlab{}.
\newblock \showarticletitle{Scalable bayesian optimization using deep neural
  networks}. In \bibinfo{booktitle}{\emph{ICML}}.
\newblock


\bibitem[\protect\citeauthoryear{Yu, Zhang, He, Chen, Xiong, and Qin}{Yu
  et~al\mbox{.}}{2018}]%
        {yu2018aesthetic}
\bibfield{author}{\bibinfo{person}{Wenhui Yu}, \bibinfo{person}{Huidi Zhang},
  \bibinfo{person}{Xiangnan He}, \bibinfo{person}{Xu Chen}, \bibinfo{person}{Li
  Xiong}, {and} \bibinfo{person}{Zheng Qin}.} \bibinfo{year}{2018}\natexlab{}.
\newblock \showarticletitle{Aesthetic-based clothing recommendation}. In
  \bibinfo{booktitle}{\emph{WWW}}.
\newblock


\bibitem[\protect\citeauthoryear{Zhang, Ai, Chen, and Croft}{Zhang
  et~al\mbox{.}}{2017}]%
        {zhang2017joint}
\bibfield{author}{\bibinfo{person}{Yongfeng Zhang}, \bibinfo{person}{Qingyao
  Ai}, \bibinfo{person}{Xu Chen}, {and} \bibinfo{person}{W~Bruce Croft}.}
  \bibinfo{year}{2017}\natexlab{}.
\newblock \showarticletitle{Joint representation learning for top-n
  recommendation with heterogeneous information sources}. In
  \bibinfo{booktitle}{\emph{CIKM}}.
\newblock


\end{thebibliography}

%

\end{document}